\documentclass[10pt,journal,cspaper,compsoc]{IEEEtran}
\usepackage{times}
\usepackage{epsfig}
\usepackage{graphicx}
\usepackage{amssymb}
\usepackage{multirow}
\usepackage{hhline}
\usepackage{booktabs}
\usepackage[fleqn]{amsmath}

\usepackage{caption}
\usepackage{algorithm2e,setspace}

\usepackage{float}
\usepackage{bm}
\usepackage{changepage}
\usepackage{subfigure}
\usepackage{multirow}
\usepackage{array}
\usepackage{todonotes}
\usepackage{changepage}

\newcolumntype{M}[1]{>{\centering\arraybackslash}m{#1}}

\ifCLASSOPTIONcompsoc
\else
\fi

\ifCLASSINFOpdf
\else
\fi

\begin{document}
\title{Weakly-Supervised Image Annotation and Segmentation with Objects and Attributes}
\author{Zhiyuan Shi, Yongxin Yang, Timothy M. Hospedales, Tao Xiang}
\IEEEcompsoctitleabstractindextext{%

 %({\color{red}to change})

%%%%%%%%%%%%%%%           ABSTRACT     %%%%%%%%%%%%%%%%%%%%%%%

\begin{abstract}
We propose to model complex visual scenes using a non-parametric Bayesian model  learned from weakly labelled images abundant on media sharing sites such as Flickr. Given weak image-level annotations of objects and attributes without  locations or associations between them, our model aims to learn  the appearance of object and attribute classes as well as their association on each object instance. Once learned, given an image, our model can be deployed to tackle a number of vision problems in a joint and coherent manner, including recognising objects in the scene (automatic object annotation), describing objects using their attributes (attribute prediction and association), and localising and delineating the objects (object detection and semantic segmentation). This is achieved by developing a novel Weakly Supervised Markov Random Field Stacked Indian Buffet Process (WS-MRF-SIBP)  that models objects and attributes as latent factors and explicitly captures their correlations within and across superpixels.   Extensive experiments on benchmark datasets demonstrate that our weakly supervised model significantly outperforms  weakly supervised alternatives and is often  comparable with existing strongly supervised models on a variety of tasks including semantic segmentation, automatic image annotation and retrieval based on object-attribute associations. 
\end{abstract}

% not want either math or citations to appear in the abstract.

% Note that keywords are not normally used for peer review papers.
}

% make the title area%
\maketitle

\section{Introduction}

One of the many incredible features of the human visual system is that it is able to generate rich description of scene content after a glance at an image. Such a description typically contains nouns and adjectives, corresponding to objects and their associated attributes, respectively. For example, an image can be described as containing ``a person in red and a shiny car''. In addition, humans can effortlessly delineate each object in the scene. One of the key objectives of computer vision research in the past five decades is to imitate this ability, resulting in intensive studies of a number of  fundamental computer vision problems including recognising objects in the scene (object annotation\footnote{\textcolor{black}{Note that while annotation sometimes refers to human created ground truth, here it refers to automatic tagging of an image with detected object categories \cite{Feng_2013_ICCV,LiSocherFeiFei2009}.}}) \cite{lsvm_pami,pascalvoc2007}, describing the objects using their attributes (attribute prediction and association) \cite{Farhadi09CVPR,kulkarni2011descriptions,lampert13AwAPAMI}, and localising and delineating the objects (object detection and semantic segmentation) \cite{singh_cvpr2013,Jimei_cvpr2014,Tighe_IJCV2015}.  

\begin{figure}[t!]
\centering
\includegraphics[width=\linewidth]{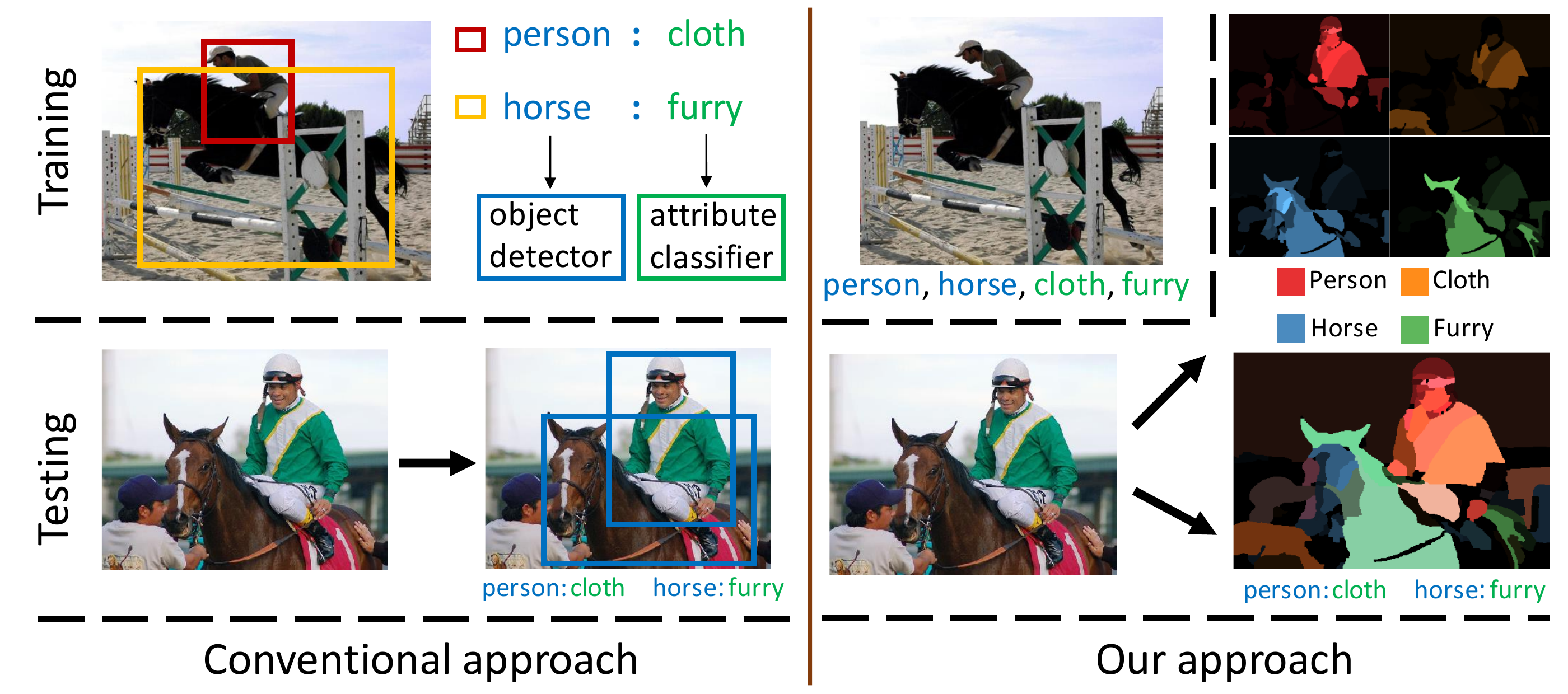}

\caption{Comparing our weakly supervised approach to  object-attribute association learning to the conventional strongly supervised approach.}
\label{fig:schematic}
\end{figure}

Although all of these problems are closely related, existing studies typically focus on one problem only, or two of them but solve them independently. Additionally, most studies employ fully supervised models learned from strongly labelled data.  Specifically, in a conventional supervised approach (Fig.~\ref{fig:schematic}) images are strongly labelled with object bounding boxes or segmentation masks, and associated attributes, from which object detectors and attribute classifiers are learned. Given a new image,  the learned object detectors are first applied to find object locations, where the attribute classifiers are then applied to produce the object descriptions. However, this conventional fully supervised and independent learning approach has two critical limitations: (1) Considering there are over 30,000 object classes distinguishable to humans \cite{Biederman_1987}, an large number of attributes to describe them, and a much larger number of object-attribute combinations, fully supervised learning is not scalable due to the lack of fully labelled training data. (2) Tackling closely related tasks jointly in a single model can be beneficial.  In particular, recent studies have shown that modelling attributes boosts object prediction accuracy and vice-versa \cite{Zheng_cvpr2014}, and localising objects helps both automatic object annotation and attribute prediction \cite{Zhenyang_eccv14}. 

We aim to overcome these limitations by solving all of these problems jointly with an object+attribute model learned from weakly labelled data, i.e.,  images with object and attribute labels but neither their associations nor their locations in the form of  bounding boxes or segmentation masks (see Fig.~\ref{fig:schematic}). As such weakly labelled images are abundant on media sharing websites such as Flickr. Therefore lack of training data would never be a problem. 

However, learning strong semantics, i.e.~explicit object-attribute association for each object instance  from weakly labelled images is extremely challenging due to the label ambiguity: a real-world image with the tags ``dog, white, coat, furry'' could contain a furry dog and a white coat or a furry coat and a white dog. Furthermore, the tags/labels typically only describe the foreground/objects. There could be a white building in the background which is ignored by the annotator, and a computer vision model must infer that this is not what the tag `white' refers to. A desirable model thus needs to jointly learn multiple objects, attributes and background clutter in a single framework in order to explain away ambiguities in each by knowledge of the other. In addition,  a potentially unlimited number of attributes can co-exist on a single object (e.g.~there are hundreds of different ways to describe the appearance of a person) which are almost certainly not labelled exhaustively. % in each training image. 
They also need to be modelled so that they do not act as distractors that have a detrimental effect on the understanding of objects and attributes of interest. For instance, annotators  may label bananas in training images but  not bother to label yellow. Even if yellow has never been used as an attribute in the training set, the model should be able to infer yellow as a latent attribute \cite{fu2013latentAttrib} and associate it with the bananas,  so that other colours would not be assigned wrongly.  %As reviewed in details in Section \ref{sec:related work}, none of the exist models is suitable. 

To this end, we develop a novel unified framework capable of jointly learning objects, attributes and their associations. The framework is illustrated schematically in Fig.~\ref{fig:schematic}, where weak annotations in the form of a mixture of objects and attributes are transformed into object and attribute associations with object segmentation. Under the framework, given a training image with image level labels of objects and attributes, the image is first over-segmented into superpixels; the joint object and attribute annotation  and segmentation problem thus becomes a superpixel multi-label classification problem whereby each superpixel can have one object label but an arbitrary number of attribute labels. Treating each label as a factor, we develop a novel factor analysis solution by  generalising the non-parametric Indian Buffet Process (IBP) \cite{Griffiths_2011_JMLR}. 

The IBP is chosen because it is designed for explaining multiple factors that simultaneously co-exist to account for the appearance of a particular image or superpixel, e.g., such factors can be an object and its particular texture and colour attributes. Importantly, as an infinite factor model, it can automatically discover and model latent factors not defined by the provided training data labels, corresponding to latent object/attributes as well as structured background `stuff' (e.g.~sky, road). However, the conventional IBP is limited in that it is unsupervised and, as a flat model, applies to either superpixels or images, but not both; it thus cannot be directly applied to our problem. Furthermore, the standard IBP is unable to exploit cues critical for segmentation and object-attribute association by modelling the correlation of factors within and across superpixels in each image. Such within-superpixel correlation captures the co-occurrence relations such as cars are typically metal and bananas are typically yellow, whilst the across-superpixel correlation dictates that neighbouring superpixels are likely to have similar labels.  To overcome these limitations, we formulate a novel variant of IBP, termed Weakly Supervised Markov Random Field Stacked Indian Buffet Process (WS-MRF-SIBP). It differs from the conventional IBP in the following ways: (1) By introducing hierarchy into IBP, WS-MRF-SIBP is able to explain images as groups of superpixels, each of which has an inferred multi-label description vector corresponding to an object and its associated attributes. (2) It learns from weak image-level supervision, which is disambiguated into multi-label superpixel explanations. (3)  Two types of Markov Random Field (MRF) over the hidden factors of an image are introduced to the model correlations: across-superpixel MRF to exploit spatial smoothness and within-superpixel MRF to exploit co-occurrence statistics of different attributes with objects. 

\section{Related work}
\label{sec:related work}

Our work is related to a wide range of computer vision problems including image classification, object recognition, attribute learning, and semantic segmentation. It is thus beyond the scope of this paper to present a comprehensive review. Since our approach differs from most existing ones in that it attempts to address all of these problems jointly using a single model learned from weakly labelled data, we  shall focus on reviewing studies that solve multiple problems simultaneously and/or use a weakly supervised learning approach.

\noindent \textbf{Learning object-attribute associations}\quad   
Attributes have been used to describe objects \cite{Farhadi09CVPR,Turakhia_2013_ICCV}, people \cite{bourdev2011attribposelet}, clothing \cite{Chen_2013_ECCV,Berg_eccv_2010}, scenes \cite{wang2013wslAttrLoc}, faces \cite{siddiquie2011img_attrib_query,Kumar_tpami_2011}, and video events \cite{fu2013latentAttrib}.  However, most previous studies learn and infer object and attribute models separately, e.g., by independently training binary classifiers, and require strong annotations/labels indicating object/attribute locations and/or associations if the image is not dominated by a single object.  A few recent studies have learned object-attribute association explicitly \cite{wang2009attrib_class_sal,Kovashka_2011_ICCV,kulkarni2011descriptions,wang2013wslAttrLoc,Wang_2013_ICCV,Wang_2010_ECCV,Mahajan_2011_ICCV,Justin_cvpr_2015}. Different from our approach, \cite{wang2009attrib_class_sal,Wang_2013_ICCV,Wang_2010_ECCV,Mahajan_2011_ICCV} only train and test on unambiguous data, i.e.~images containing a single dominant object, assuming object-attribute association is known at training;  moreover, they allocate exactly one attribute per object.  Kulkarni et al.~\cite{kulkarni2011descriptions} model the more challenging PASCAL VOC type of data with multiple objects and attributes co-existing. However, their model is pre-trained on object and attribute detectors learned using strongly annotated images with object bounding boxes provided. The work in \cite{wang2013wslAttrLoc} also does object segmentation and object-attribute prediction. But its model is learned from strongly labelled images in that object-attribute association are given during training; and importantly prediction is restricted to object-attribute pairs seen during training. In summary no existing studies learn flexible object-attribute association from weakly labelled data as we do in this work.

Some existing studies aim to perform attribute-based query \cite{Rastegari_CVPR13,Kovashka_2013_ICCV,multiattrs_cvpr2012,siddiquie2011img_attrib_query}. In particular, recent studies  have considered how to calibrate \cite{multiattrs_cvpr2012} and fuse \cite{Rastegari_CVPR13} multiple attribute scores in a single query. We go beyond these by supporting object+multi-attribute conjunction queries. Moreover, existing methods either require bounding boxes or assume simple data with single dominant objects, and do not reason jointly about multiple attribute-object associations. This means that they would be intrinsically challenged in reasoning about (multi)-attribute-object queries on challenging data with multiple objects and multiple attributes in each image (e.g., querying furry brown horse, in a dataset with black horses and furry dogs in the same image). In other words, they cannot be directly extended to solve query by object-attribute association.

\noindent \textbf{Weakly supervised semantic segmentation}\quad  Most existing semantic segmentation models are fully supervised requiring pixel-level labels \cite{singh_cvpr2013,Jimei_cvpr2014,Tighe_IJCV2015,Zheng_cvpr2014}.  A few weakly supervised semantic segmentation methods have been presented recently, exploring a variety of models such as conditional random fields (CRF) \cite{xu_cvpr2014,vezhnevets_cvpr2012}, label propagation \cite{Rubinstein12Annotation} and clustering \cite{Yang_cvpr2013}. \textcolor{black}{More recently, convolutional neural networks have been shown to work very well for this challenging task, either in a fully supervised fashion \cite{long_cvpr_2015,chen_iclr_2015} or a weakly supervised fashion \cite{Pathak_iclr_2015,Jia_cvpr_2015}. However, these methods require a large-scale annotated dataset (e.g. ImageNet) to train or pre-train a deep CNN model for feature representation.}   Another closely related problem is two- or multi-class co-segmentation \cite{rubenstein2013unsupJointDisc}, where the task is to segment shared objects from a set of images. Although co-segmentation does not require image labels \emph{per-se}, it indeed assumes common objects across multiple training images. Like previous semantic segmentation methods, we focus on how to learn a model to segment unseen and unlabelled test images, rather than solely segmenting the training images as in co-segmentation. Importantly, all of these previous methods only focus on object labels (nouns). Our method provides a mechanism to jointly learn objects, attributes and their associations (adjective-noun pairs). We show that attribute labels  provide valuable complementary information via inter-label correlation, especially under this more ambitious weakly supervised setting. 

This work is not the first to exploit the benefit of joint modelling object and attributes for segmentation. Recently, Zheng et al \cite{Zheng_cvpr2014} formulated joint visual attribute and object segmentation as a multi-label problem using a fully connected CRF. Similarly, a model was proposed in \cite{Zhenyang_eccv14} to learn and extract attributes from segmented objects, which notably improves object classification accuracy. However, both of these methods are fully supervised, requiring pixel-level ground truth for training. In contrast, our proposed approach can cope with weakly labelled data to alleviate the burden of strong annotation.

{\noindent \textbf{Weakly supervised learning: our model vs.~discriminative models}\quad
Discriminative methods underpin many high performance  recognition and annotation studies \cite{Farhadi09CVPR,RussakovskyECCV10,bourdev2011attribposelet,wang2013wslAttrLoc,sadeghi2011recoPhrases,kulkarni2011descriptions}. 
Similarly existing weakly supervised learning (WSL) methods are also dominated by discriminative models. Apart from the  mentioned conditional random field (CRF) \cite{xu_cvpr2014,vezhnevets_cvpr2012}, label propagation \cite{Rubinstein12Annotation} and clustering \cite{Yang_cvpr2013} models, some  discriminative multi-instance learning (MIL)  models were proposed \cite{nguyen2010svm_miml,deselaers2012wslGeneric}. Our model is a  probabilistic generative model.   Compared to a discriminative model, the strengths of a generative model for WSL is its abilities to infer latent factors corresponding to background clutter and un-annotated  objects/attributes, and to model them jointly in a single model so as to explain away the ambiguity existing in the weak image-level labels. Very recently deep learning based  image captioning has started to attract attention \cite{Vinyals_2015_CVPR,Karpathy_2015_CVPR}. Generating a natural sentence describing the  an image is  a harder task than listing nouns and adjectives - other words including verbs (action) and prepositions (where) need to be inferred and language syntax needs to be followed in the generated text description. However, these neural network models are essentially still discriminative models and have the same drawbacks as other discriminative models for WSL.

\noindent \textbf{Weakly supervised learning: our model vs.~other probabilistic generative models}\quad 
The flexibility of generative probabilistic models and their suitability particularly for WSL have seen them successfully applied to a variety of WSL tasks \cite{LiSocherFeiFei2009,fu2013latentAttrib,socher2010sslModalities,Shi_2013_ICCV}.  These studies often generalise  probabilistic topic models (PTM) \cite{blei2003lda}. However PTMs are limited for explaining objects and attributes in that latent topics are competitive - the fundamental assumption is that an object is a horse \emph{or} brown \emph{or} furry. They intrinsically do not account for the reality that it is \emph{all} at once.  
In contrast, our model generalises Indian Buffet Process (IBP) \cite{velez2009variational,Griffiths_2011_JMLR}. The IBP is a latent feature model  that can independently activate each latent factor, explaining imagery as a weighted sum of active factor appearances. 

Our Weakly Supervised Markov Random Field Stacked Indian Buffet Process (WS-MRF-SIBP) differs significantly from the standard flat and unsupervised IBP in that it is hierarchical  to model grouped data (images composed of superpixels) and weakly supervised. This allows us to exploit image-level weak supervision, but disambiguate it to determine the best explanation in terms of which superpixels correspond to un-annotated background, which superpixels correspond to which annotated objects, and which objects have which attributes. In addition, a Markov random field (MRF) is integrated into the IBP to model correlations of factors both within and across superpixels. A few previous studies \cite{Verbeek2007,Zhao_eccv_2010,Bolei_cvpr11} generalise  classic PTMs \cite{blei2003lda} by integrating a MRF to enforce  spatial coherence across topic labels of neighbouring regions. Unlike these methods, we generalise the IBP by defining the MRF over hidden factors. Furthermore, beyond spatial coherence we also define a factorial MRF to capture attribute-attribute and attribute-object co-occurrences within superpixels.

\noindent \textbf{Our contributions} \quad This paper makes the following key contributions: (i) We for the first time jointly learn all object, attribute and background appearances, object-attribute association, and their locations from realistic weakly labelled images including multiple objects with cluttered background; (ii) we formulate a novel weakly supervised Bayesian model by generalising the IBP to make it weakly supervised, hierarchical, and integrate two types of hidden factor MRFs to learn and exploit spatial coherence and factor co-occurrence; (iii) Once learned from weakly labelled data, our model can be deployed for various tasks including semantic segmentation, image description and image query, many of which rely on predicting strong object-attribute association. Extensive experiments on benchmark datasets demonstrate that on all tasks our model  significantly outperforms a number of weakly supervised baselines and in many cases is comparable to  strongly supervised alternatives. A preliminary version of our work was described in \cite{Shi_2014_ECCV}.

%\noindent \textbf{Contributions} \quad In this paper we make three key contributions: (i) We for the first time jointly learn all object, attribute and background appearances, object-attribute association, and their locations from realistic weakly labelled images; (ii) We formulate a novel  weakly supervised non-parametric Bayesian model by generalising the Indian Buffet Process; (iii) We demonstrate various image description and query tasks, including challenging tasks relying on predicting strong object-attribute association. Extensive experiments on benchmark datasets demonstrate that our model is comparable to strongly supervised alternatives and significantly outperforms a number of weakly supervised baselines.

\section{Methodology}

\subsection{Image representation}
\label{sec:representation}
Given a set of images labelled with image-level object and attribute labels, but without explicitly specifying which attribute is associated with which object, we aim to learn a model that, given a  new image,  segments each object in the image and assign both object and attribute labels to it. As in most previous semantic segmentation works \cite{Tighe_IJCV2015,singh_cvpr2013,Jimei_cvpr2014,xu_cvpr2014,vezhnevets_cvpr2012}, we first decompose each image into superpixels which are over-segmented image patches that typically contain object parts. The problem of joint object and attribute annotation and segmentation thus boils down to multi-label classification of each superpixel, from which various tasks such as automatic image-level annotation, object-attribute association, and object segmentation can be performed. 

Each image $i$ in a training set is decomposed into $N_i$ superpixels using a recent hierarchical  segmentation algorithm \cite{amfm_pami2011}\footnote{ We set the segmentation threshold to $0.1$ to obtain a single over-segmentation from the hierarchical segmentations for each image.}. Each segmented superpixel is represented using two normalised histogram features: SIFT and Color. (1) SIFT: we extract regular grid (every 5 pixels) colorSIFT \cite{vandeSandeTPAMI2010} at four scales. A 256 component GMM  model is constructed on the collection of ColourSIFTs from all images. We compute Fisher Vector + PCA for all regular points in each superpixel following \cite{Aggregating_TPAMI_2011}. The resulting reduced descriptor is 512-D for every segmented region. (2) Colour: We convert the image to quantised LAB space 8$\times$8$\times$8. A 512-D color histogram is then computed for each superpixel. The final normalised 1024-D feature vector concatenates SIFT and Colour features together.

\begin{figure}[t]
\centering
\subfigure[WS-SIBP]{
  \centering
  \includegraphics[height=.47\linewidth]{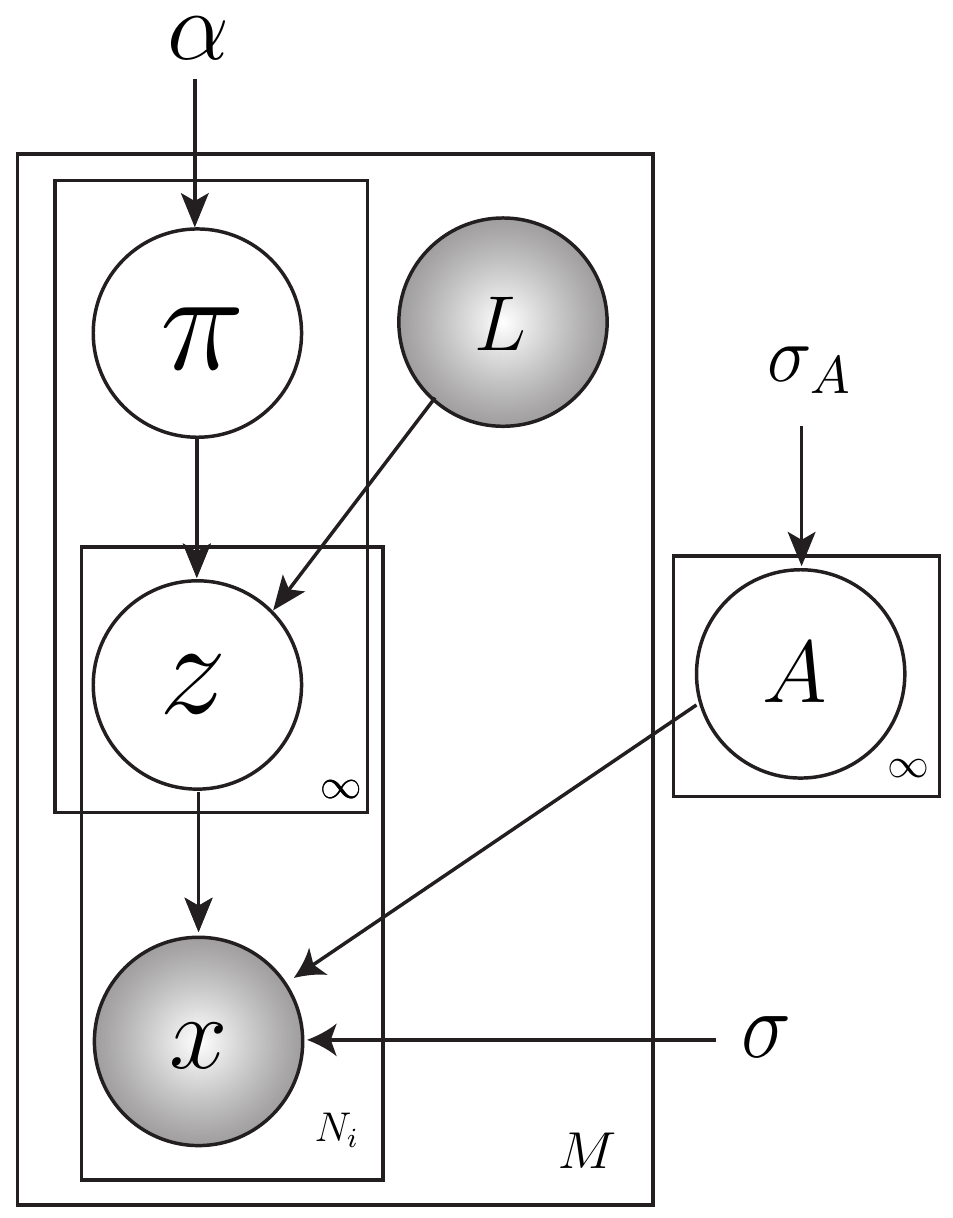}
  %\caption{aPascal}
  \label{fig:m_imagelevel}
}%
\subfigure[WS-MRF-SIBP]{
  \centering
  \includegraphics[height=.47\linewidth]{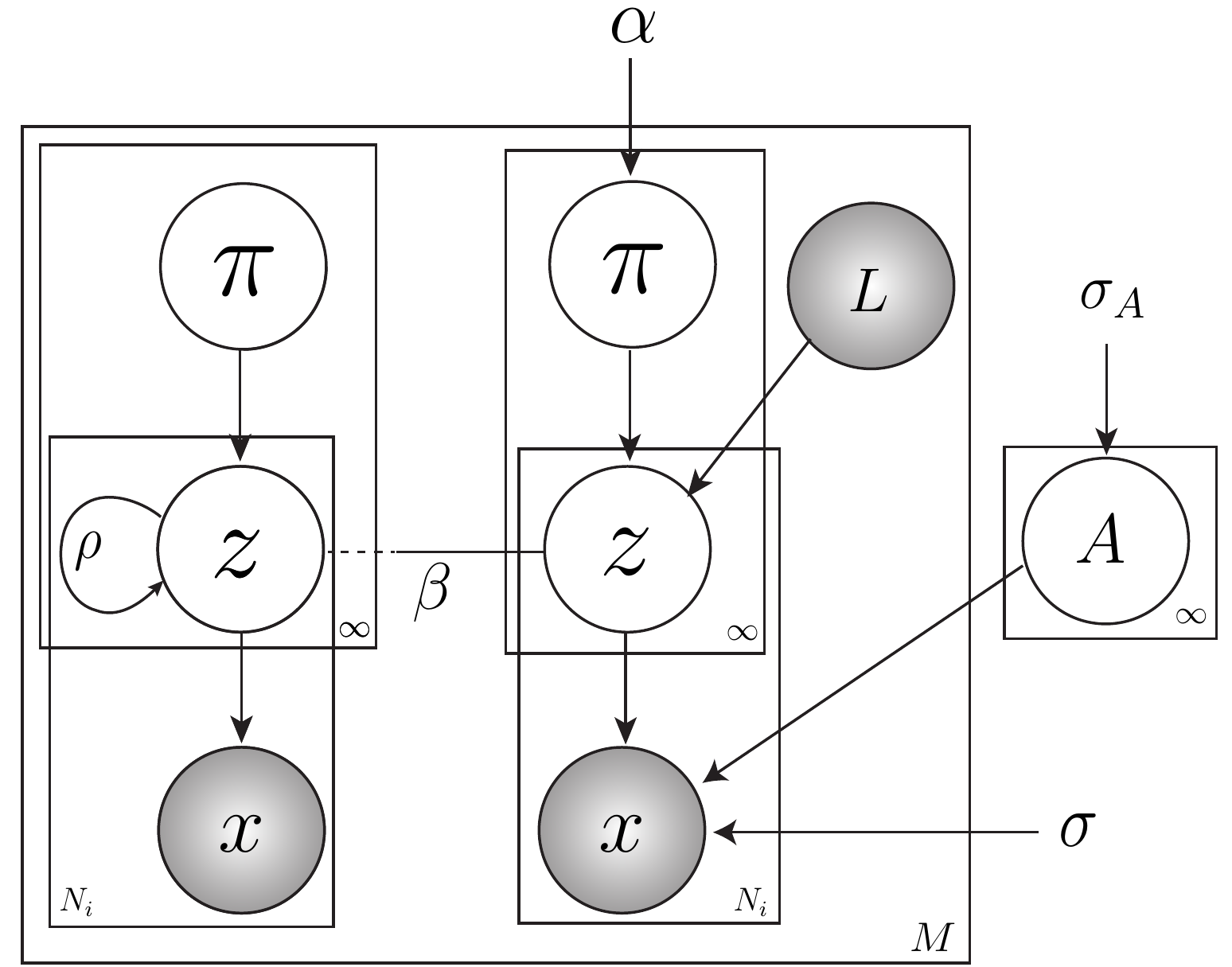}
  %\caption{aPascal}
  \label{fig:m_infer}
}%
\caption{The probabilistic graphical models representing our WS-SIBP and WS-MRF-SIBP. Shaded nodes are observed.}
\label{fig:PGM}
\end{figure}

\subsection{Model formulation}
We propose a non-parametric Bayesian model that learns to describe images composed of superpixels from weak image-level object and attribute annotation. In our model, each superpixel is associated with an infinite latent factor vector indicating if it corresponds to (an unlimited variety of) unannotated background clutter, or an object of interest, and what set of  attributes are possessed by the object.  Given a set of images with weak labels and segmented into superpixels, we need to learn: (i) which are the corresponding superpixels shared by all images with a particular label, (ii) which superpixels correspond to unannotated background, and (iii) what is the appearance of each object, attribute and background type. Moreover, since multiple labels (attribute and object) can apply to a single superpixel, we need to disambiguate which aspects of the appearance of each superpixel are due to each of the (unknown) associated object and attribute labels.  To address the weakly-supervised learning tasks we build on the IBP \cite{velez2009variational} and  introduce in Sec.~\ref{sec:WS-SIBP} a weakly-supervised stacked Indian Buffet process (WS-SIBP) to model data represented as bags (images) of instances (superpixels) with bag-level labels (image-level annotations). This is analogous to the notion of documents in topic models \cite{blei2003lda}. Furthermore, to fully exploit spatial and inter-factor correlation, two types of MRFs are integrated (see Sec.~\ref{sec:WS-MRF-IBP}), resulting in the full model termed  WS-MRF-SIBP\footnote{The codes for both models will be made available at http://zhiyshi.com/publication/}. 

\subsubsection{WS-SIBP}
\label{sec:WS-SIBP} 
We aim to associate each image/superpixel with a latent factor vector whose elements will correspond to objects, attributes and/or unannotated attribute/background present in that image/superpixel. Let  images $i$ be represented as bags of superpixels $\mathbf{X}^{(i)}=\{\mathbf{X}^{(i)}_{j\cdot}\}$, where  the notation $\bm{X}_{j\cdot}$ means the vector of row $j$ in matrix $\bm{X}$, i.e.~the 1024-D feature vector representing the $j$-th superpixel, and $j\in\{1\dots N_i\}$. Assuming there are $K_o$ object categories and $K_a$ attributes in the provided image-level annotations, they are represented by the first $K_{oa}=K_o+K_a$ latent factors. In addition, an unbounded number of further factors are available to explain away background clutter in the data, as well as discover unannotated latent attributes. At training time, we assume a 
 binary label vector $L^{(i)}$ for objects and attributes is provided for each image $i$. So $L^{(i)}_k=1$ if attribute/object $k$ is present, and zero otherwise. Also $L^{(i)}_k =1$ for all $k>K_{oa}$. That is, without any labels, we assume all background/latent attribute types can be present. With these assumptions, the generative process (see Fig.~\ref{fig:PGM}(a)) for the image $i$ is as follows:
\\~\\
\noindent For each latent factor $k\in1\dots \infty$:

\begin{enumerate}

\item Draw an appearance distribution mean $\mathbf{A}_{k\cdot} \sim \mathcal{N}(0,\sigma_{A}^2\bm{I})$.

\end{enumerate}
For each image $i\in1\dots M$:
\begin{enumerate}
\item Draw a sequence of i.i.d. random variables $v_1^{(i)},v_2^{(i)} \dots \sim \mbox{Beta}(\alpha, 1)$,

\item Construct an image prior $\pi_k^{(i)} = \prod\limits\limits_{t=1}^{k}v_t^{(i)}$,

\item Input weak annotation $L_k^{(i)} \in \{0, 1\}$,

\item For each superpixel  $j\in1\dots N_i$:

\begin{enumerate}

\item Sample state of each latent factor $k$: $z_{jk}^{(i)} \sim \mbox{Bern}(\pi_k^{(i)}L_k^{(i)})$,
\item Sample superpixel appearance: $\bm{X}_{j\cdot}^{(i)} \sim \mathcal{N}(\bm{Z}_{j\cdot}^{(i)}\bm{A},\sigma^2\bm{I}) $.

\end{enumerate}

\end{enumerate}

\noindent where $\mathcal{N}$, $\mbox{Bern}$ and $\mbox{Beta}$ respectively correspond to Normal, Bernoulli and Beta distributions with the specified parameters. The Beta-Bernoulli and Normal-Normal conjugacy are chosen because they allow more efficient inference. $\alpha$ is the prior expected sparsity of annotations and $\sigma^2$ is the prior variance in appearance for each factor. 

This generative process encodes the assumptions that the available factors for each superpixel are determined by the image level labels if given (generative model for $\mathbf{Z}$); and that multiple factors come together to explain each superpixel (generative model for $\mathbf{X}$ given $\mathbf{Z}$).

\noindent\textbf{Joint probability:}
Denote hidden variables by $\bm{H} = \{\bm{\pi}^{(1)},\dots,\bm{\pi}^{(M)}, \bm{Z}^{(1)},\dots, \bm{Z}^{(M)}, \bm{A}\}$, $M$ images in a training set by $\bm{X} = \{\bm{X}^{(1)},\dots,\bm{X}^{(M)}\}$, %and parameters by $\bm{\Theta}=\{\alpha, \sigma_A, \sigma, \bm{{L}},\beta,\rho\}$. Then 
and parameters by $\bm{\Theta}=\{\alpha, \sigma_A, \sigma, \bm{L}\}$. Then the joint  probability of the variables and data given the parameters is:
\begin{align}
p(\bm{H}, \bm{X}|&\bm{\Theta})=\nonumber \\
&\prod_{i=1}^{M}\bigg(\prod_{k=1}^{\infty}\Big(p(\pi_k^{(i)}|\alpha)\prod_{j=1}^{N_i}p(z_{jk}^{(i)}|\pi_k^{(i)}, L_k^{(i)})\Big)\nonumber \\
%&\prod_{i=1}^{M}\bigg(\prod_{k=1}^{\infty}\Big(p(\pi_k^{(i)}|\alpha)\prod_{j=1}^{N_i}p(z_{jk}^{(i)}|\pi_k^{(i)}, L_k^{(i)},\beta,\rho)\Big)\nonumber \\
&\cdot\prod_{j=1}^{N_i}p(\bm{X}_{j\cdot}^{(i)}|\bm{Z}^{(i)}_{j\cdot}, \bm{A}, \sigma)\bigg) \nonumber \\
&\cdot\prod_{k=1}^{\infty}p(\mathbf{A}_{k\cdot}|\sigma_{A}^2)\label{eq:joint}.
\end{align}

\noindent Learning our model (detailed in Sec.~\ref{sec:modellearn}) aims to compute the posterior $ p(\bm{H} | \bm{X}, \bm{\Theta})$ for: disambiguating and localising all the annotated ($L^{(i)}$) objects and attributes among the superpixels (inferring $\bm{Z}^{(i)}_{j\cdot}$), inferring the attribute and background prior for each image (inferring $\bm{\pi}^{(i)}$), and learning the appearance of each factor (inferring $\mathbf{A}_{k\cdot}$).

\subsubsection{WS-MRF-SIBP}
\label{sec:WS-MRF-IBP}
Now we generalise the WS-SIBP to WS-MRF-SIBP by introducing two types of factor correlation.\\
\noindent\textbf{Spatial MRF across superpixels:}
Each superpixel's latent factors are so far drawn from the image prior $\pi_{k}^{i}$ -- independently of their neighbours (Eq.~(\ref{eq:joint})). Thus  spatial structure is ignored in WS-SIBP, even though adjacent superpixels are  strongly correlated in real images \cite{Verbeek2007}. Inspired by the successful use of random fields for capturing the spatial coherence of image region labels \cite{Zhao_eccv_2010,Verbeek2007,Bolei_cvpr11}, we introduce a MRF with connections between adjacent nodes (superpixels). Specifically, the following MRF potential \cite{Verbeek2007, Bishop_2006} is introduced to the generative process for $\mathbf{Z}$ to correlate the  superpixel factors drawn in image $i$ spatially:
%we build a Markov Random Field over hidden topics. The topic random field, depicted as a generative model in Figure., introduces explicit couplings between the labels of adjacent regions in an image. This allows the TRF model the ability to capture local correlations that would be missed under the conditional independence assumption of spatial LDA. placing a MRF prior on hidden topic labels zd. 
\begin{eqnarray}
\Phi(\bm{Z}^{(i)}_{\cdot k}) = \exp \sum_{j,m \in N_i}^{ } \beta \mathbf{I} (z^{(i)}_{jk}=z^{(i)}_{mk}),
%p(z^{i}|\beta)=p(\{z^{i}_{1},...,z^{i}_{N_{i}} \}|\beta) \propto \exp \sum_{j\sim m}^{ } \beta I (z^{i}_{j}=z^{ik}_{mk})) 
\end{eqnarray}
where $j,m\in N_i$ enumerates node pairs that are  neighbours in image $i$. The indicator function I returns one when its argument is true, i.e., when neighbouring superpixels have the same assignment for factor $k$. $\beta$ is the coupling strength parameter of the MRF, which controls how likely they have the same label \emph{a priori}. The initial WS-SIBP formulation can be obtained by setting $\beta=0$. The spatial MRF is encoded for all given $K_{oa}$ and newly discovered factors.

\noindent\textbf{Factorial MRF within superpixel:} Although individual factors are now correlated spatially, we do not yet model any inter-factor co-occurrence statistics within a single superpixel (as in most other MRF applications \cite{Verbeek2007,Zhao_eccv_2010}). However, exploiting this information (e.g., \emph{person} superpixels more likely to share attribute \emph{clothing}, than \emph{metallic}) is important, especially in the ambiguous WSL setting. To represent these inter-factor correlations, we introduce a factorial MRF via the following potential on $\mathbf{Z}$: 
%\begin{equation}
%%\Psi(z^{i}_{jk},z^{i}_{jn}) = \left \{  \begin{array}{l l}
%%  0 & \quad \text{if $k=n$ }\\
%%  \rho\mathcal{M} (k,n) & \quad \text{otherwise,}
%  \Psi(z^{i}_{jk},z^{i}_{jn}) = \left \{  \begin{array}{l l}
%  0 & \quad \text{if $k=n$ }\\
%  \rho\mathcal{M} (k,n) & \quad \text{otherwise,}
%\end{array} \right.
%\end{equation} 
\begin{eqnarray}
\Psi(\mathbf{Z}_{j\cdot}^{(i)}) &=& \exp \sum^{\infty}_{k,l}\psi(z^{(i)}_{jk},z^{(i)}_{jl}) \\
\psi(z^{(i)}_{jk},z^{(i)}_{jl}) & = &  \left \{  \begin{array}{l l}
  0 & \quad \text{if $k=l$ }\\
  \rho \mathcal{M}_{kl} & \quad \text{otherwise,}
\end{array} \right.
\end{eqnarray} 
\noindent where $\rho$ controls the importance of the factorial MRF, and $\mathcal{M}_{kl}$ is an element of the factor correlation matrix $\mathbf{M}$ that encodes the correlation between factor $k$ and factor $l$. In the traditional strongly-supervised scenario, $\mathbf{M}$ can be trivially learnt from the fully labelled annotations. In the WSL scenario, $\mathbf{M}$ cannot be determined directly. We will discuss how to learn $\mathbf{M}$ in Sec.~\ref{sec:modellearn}. % Note that $\mathbf{M}$ encodes correlation on various factors depending on the real world data. If attribute annotation is available, we consider the first $K_{oa}$ latent factors. Otherwise, we allow both foreground and background factors to be captured, thus discovering the latent attributes. 

%Here we initialize the $\mathbf{M}$ as a correlation matrix (see Fig.~\ref{fig:m_imagelevel}) using image-level tag which we only have in weakly supervised setting. As shown in Fig.~\ref{fig:m_imagelevel}, the initial $\mathbf{M}$ is very noisy compare to the ground true correlation (see Fig.~\ref{fig:m_pixellevel}) which generated from pixel-level annotation. In order to disambiguate the initial relations, we iteratively refine the $\mathbf{M}$ with the inferred factors in M step (see Sec.~\ref{sec:modellearn}). The correlation is much cleaner in the updated $\mathbf{M}$ (see Fig.~\ref{fig:m_infer}).}

%Denote hidden variables by $\bm{H} = \{\bm{\pi}^{(1)},\dots,\bm{\pi}^{(M)}, \bm{Z}^{(1)},\dots, \bm{Z}^{(M)}, \bm{A}\}$, observed images by $\bm{X} = \{\bm{X}^{(1)},\dots,\bm{X}^{(M)}\}$, and model parameters by $\bm{\Theta}=\{\alpha, \sigma_A, \sigma, \bm{{L}}\}$. Then the joint  probability of the variables and data given the parameters is:

\noindent\textbf{WS-MRF-IBP Prior:} Overall, combining the two  MRFs, the latent factor prior 
\begin{align}
p(\bm{Z}^{(i)}|\bm{\pi}^{(i)},L^{(i)})=\prod_{k=1}^{\infty}\prod_{j=1}^{N_i}p(z_{jk}^{(i)}|\pi_k^{(i)}, L_k^{(i)})\nonumber
\end{align}
used by Eq.~(\ref{eq:joint}), is now replaced by:
\begin{flalign*}
%\begin{align}
&p(\bm{Z}^{(i)}|\bm{\pi}^{(i)},L^{(i)},\beta,\rho) \propto \exp\Big(\sum_{k=1}^{\infty}\sum_{j=1}^{N_i}\log p(z_{jk}^{(i)}|\pi_k^{(i)}, L_k^{(i)}) && \nonumber \\
& + \sum_{k=1}^{\infty} \log\Phi(\bm{Z}^{(i)}_{\cdot k})+\sum_{j=1}^{N_i}\log\Psi(\bm{Z}^{(i)}_{j \cdot})\Big). &&
\end{flalign*}
%\end{align}

\noindent and the list of model parameters $\bm{\Theta}$ is extended to $\bm{\Theta}=\{\alpha, \sigma_A, \sigma, \bm{L}, \rho, \beta, \mathbf{M}\}$. 

\subsection{Model learning}
\label{sec:modellearn}

%\SetAlgorithmName{}{}{}
%\renewcommand{\thealgocf}{} 
%\SetAlgoCaptionSeparator{}
%\LinesNumbered
\begin{algorithm*}[t]
\small
\setstretch{0.5}
\SetAlgoLined
\While{not converge}
{
\For {k = 1 \emph{\KwTo} $K_{max}$}{
%\setstretch{2}

\begin{flalign}
\bm{\phi}_k =&(\frac{1}{\sigma^2}\sum\limits_{i=1}^{M}\sum\limits_{j=1}^{N_i}\nu_{jk}^{(i)}(\bm{X}_{j\cdot}^{(i)}-\sum\limits_{l:l \neq k} \nu_{jl}^{(i)}\bm{\phi}_l)) \cdot(\frac{1}{\sigma_A^2}+\frac{1}{\sigma^2}\sum\limits_{i=1}^{M}\sum\limits_{j=1}^{N_i}\nu_{jk}^{(i)})^{-1} \label{eq:update_a}  \\
\bm{\Phi}_k =& \Big(\frac{1}{\sigma_A^2}+\frac{1}{\sigma^2}\sum\limits_{i=1}^{M}\sum\limits_{j=1}^{N_i}\nu_{jk}^{(i)}\Big)^{-1}\bm{I} \label{eq:update_a2} \qquad \triangleright \textrm{Update appearance model including mean and covariance.} 
\end{flalign}
}

\For{i = 1 \emph{\KwTo} $M$}
{
\For {k = 1 \emph{\KwTo} $K_{max}$}{
%\setstretch{1.75}
\vspace{-0cm}
\begin{flalign}
\hspace{0cm} \tau_{k1}^{(i)} =& \alpha + \sum\limits_{m=k}^{K_{max}}\sum\limits_{j=1}^{N_i}\nu_{jm}^{(i)} + \sum\limits_{m=k+1}^{K_{max}}(N_i - \sum\limits_{j=1}^{N_i}\nu_{jm}^{(i)})(\sum\limits_{s=k+1}^{m}q_{ms}^{(i)}) \\
\hspace{0cm}\tau_{k2}^{(i)} =& 1 + \sum\limits_{m=k}^{K_{max}}(N_i - \sum\limits_{j=1}^{N_i}\nu_{jm}^{(i)})q_{mk}^{(i)} \qquad \qquad \qquad \triangleright \textrm{Update image prior for every factor k.} 
\end{flalign}

\For {j = 1 \emph{\KwTo} $N_i$}{ 
\vspace{-0cm} 
\begin{flalign}
\hspace{0cm} \eta =& \sum\limits_{t=1}^{k}(\varphi(\tau_{t1}^{(i)}) -  \varphi(\tau_{t2}^{(i)})) - \mathbb{E}_{\bm{v}}[\log(1-\prod\limits_{t=1}^{k}v_{t}^{(i)})] -\frac{1}{2\sigma^2}(\mbox{tr}(\bm{\Phi}_k)+\bm{\phi}_k\bm{\phi}_k^T \nonumber \\
&-2 \bm{\phi}_k(\bm{X}_{j\cdot}^{(i)}-\sum\limits_{l:l \neq k} \nu_{jl}^{(i)}\bm{\phi}_l)^T) \qquad \qquad \qquad \qquad \triangleright \textrm{Top-down prior and bottom-up data cues.}  \\
\hspace{0cm} {\eta}' = &\eta + \sum\limits_{m\in N(j)}\beta \eta^{(i)}_{mk} + \sum\limits_{n: n \neq k}\rho \mathcal{M}_{kn} \eta_{jn} \label{eq:update_mrf}  \qquad \triangleright \textrm{Influence of the two MRFs.}  
\end{flalign}
\begin{equation}
\hspace{0cm} \nu_{jk}^{(i)} = \frac{L_k^{(i)}}{1+\exp \left [-{\eta}' \right ]} \label{eq:update_mrf2} \qquad \triangleright \textrm{Final posterior for each latent factor $z^{(i)}_{jk}$ state.}  
\end{equation}
}
}
}

\For{i = 1 \emph{\KwTo} $M$}
{
\For {j = 1 \emph{\KwTo} $N_i$}{
\begin{equation}
\hspace{0cm} \mathbf{M} = \mathbf{M} + (\eta_{j\cdot}^{i})^{T}\eta_{j\cdot}^{i} \label{eq:update_m} \qquad \triangleright \textrm{Update the intra-superpixel correlation given inferred factors.}
\end{equation} 
}
}
}
\caption{\label{alg:inference} Variational inference for learning WS-MRF-SIBP}
\end{algorithm*}

Exact inference for $ p(\bm{H} | \bm{X}, \bm{\Theta})$ in our model is intractable, so an approximate inference algorithm in the spirit of \cite{velez2009variational} is developed. The mean field variational approximation to the desired posterior $p(\bm{H} | \bm{X}, \bm{\Theta})$ is:
\begin{equation}
q(\bm{H})=\prod_{i=1}^{M}\big(q_{\bm{\tau}}(\bm{v}^{(i)})q_{\bm{\nu}}(\bm{Z}^{(i)})\big)q_{\bm{\phi}}(\bm{A})\label{eq:meanFieldVar}
\end{equation}

\noindent where $q_{\bm{\tau}}(v_k^{(i)}) = \mbox{Beta}(v_k^{(i)}; \tau_{k1}^{(i)} \tau_{k2}^{(i)})$, $q_{\bm{\nu}}(z_{jk}^{(i)}) = \mbox{Bernoulli}(z_{jk}^{(i)}; \nu_{jk}^{(i)})$, $q_{\bm{\phi}}(\bm{A}_{k\cdot}) = \mathcal{N}(\bm{A}_{k\cdot}; \bm{\phi}_k, \bm{\Phi}_k)$ and the infinite stick-breaking process for latent factors is truncated at $K_{max}$, so $\pi_k=0$ for $k>K_{max}$. %\textcolor{red}{Is the previous statement true despite the MRF change? i think so.} 
A variational message passing (VMP) strategy \cite{velez2009variational} can be used to minimise the KL divergence of Eq.~(\ref{eq:meanFieldVar}) to the true posterior. Updates are obtained by deriving integrals of the form $\ln q(\mathbf{h})=E_{\mathbf{H}\backslash\mathbf{h}}\left[\ln p(\mathbf{H},\mathbf{X})\right] + C$ for each group of hidden variables $\mathbf{h}$.
%.~\ref{}-- using a truncation approximation to the infinite latent factor distribution -- 
These result in the series of iterative updates given in Algorithm \ref{alg:inference}, where $\varphi(\cdot)$ is the digamma function; and $q^{(i)}_{ms}$ and $\mathbb{E}_{\bm{v}}[\log(1-\prod\limits_{t=1}^{k}v_{t}^{(i)})]$ are given in \cite{velez2009variational}. 

Like \cite{Zhao_eccv_2010,Verbeek2007}, the MRF influence is via Eqs.~(\ref{eq:update_mrf}) and (\ref{eq:update_mrf2}). However, while the works of \cite{Zhao_eccv_2010,Verbeek2007} only consider spatial coherence, we further model the inter-factor correlation, which we will see is very important for our weakly supervised tasks, especially in image annotation.% using learned attributes and objects factors. 

\noindent \textbf{Factor correlation learning:}\quad 
The correlation matrix $\mathbf{M}$ is non-trivial to estimate accurately in the WSL case, in contrast to the fully-supervised case where it is easy to obtain as the correlation of superpixel annotations. In the WSL case, it can only be estimated a priori from image-level tags. However, this is a very noisy estimate. For example, an image with tags \emph{furry}, \emph{horse}, \emph{metal}, \emph{car} will erroneously suggest \emph{horse-car}, \emph{furry-metal}, \emph{horse-metal} as correlations. 

To address this, we initialise $\mathbf{M}$ coarsely with image-level labels as $\mathbf{M}=\sum_{i=1}^{M}(L_k^{(i)})^T(L_k^{(i)})$, and refine it with an EM process. During learning, we re-estimate $\mathbf{M}$ at each iteration using the disambiguated superpixel-level factors inferred by the model, as in  Eq.~(\ref{eq:update_m}). Thus as the correlation estimate improves, the estimated factors become more accurate, and vice-versa. The effectiveness of this iterative learning procedure is demonstrated in the supplementary material. 

%\textcolor{green}{Here we initialize the $\mathbf{M}$ as a correlation matrix (see Fig.~\ref{fig:m_imagelevel}) using image-level tag which we only have in weakly supervised setting. As shown in Fig.~\ref{fig:m_imagelevel}, the initial $\mathbf{M}$ is very noisy compare to the ground true correlation (see Fig.~\ref{fig:m_pixellevel}) which generated from pixel-level annotation. In order to disambiguate the initial relations, we iteratively refine the $\mathbf{M}$ with the inferred factors in M step (see Sec.~\ref{sec:modellearn}). The correlation is much cleaner in the updated $\mathbf{M}$ (see Fig.~\ref{fig:m_infer}).}

\noindent\textbf{Efficiency:}\quad 
In practice, the truncation approximation means that our WS-MRF-SIBP runs with a finite number of factors $K_{max}$ which can be freely set so long as it is bigger than the number of factors needed by both annotations and background clutter ($K_{bg}$), i.e., $K_{max} > K_o+K_a+K_{bg}$\footnote{In this work, we set $K_{max}=K_o+K_a+20$.}. Despite the combinatorial nature of the object-attribute association and localisation problem, our model is of complexity $\mathcal{O}(MNDK_{max}+K_{max}^2)$ for $M$ images with $N$ superpixels, $D$ feature dimension and $K_{max}$ truncated factors.  %\textcolor{red}{:We mentioned $K_{max}$ is much greater than $K_o+K_a+K_{bg}$. In practice, we did not set a very big number. For example, We set $K_{max}=104$ (including 20 objects and 64 attributes) for aPascal dataset. Is that OK to use $K_{max}$ in complexity. People may think the complexity is very large? or maybe we define new symbol $K$=$K_o+K_a+K_{bg}$.}

%Where the correlation matrix M can be calculated from image-level attribute annotation or patch-level attribute annotation. It can also be updated by $\eta$ after each iteration.

\subsection{Inference for test data}

At testing time, the appearance of each factor $k$, now modelled by sufficient statistics $\mathcal{N}(\bm{A}_{k\cdot}; \bm{\phi}_k, \bm{\Phi}_k)$, is assumed to be known (learned from the training data), while annotations for each test image $L^{(i)}_k$ will need to be inferred. Thus Algorithm~\ref{alg:inference} still applies, but without the appearance update terms (Eqs.~(\ref{eq:update_a}) and (\ref{eq:update_a2})) and with $L^{(i)}_k=1~\forall k$, to reflect the fact that all the learned object, attribute, and background types could be present.

\subsection{Applications of the model}
\label{sec:postproc}

Given the learned model applied to test data, we can perform the following tasks. %\textcolor{blue}{Note that the first four tasks are non-transductive, so factor appearances $(\bm{\phi}_k, \bm{\Phi}_k)$  are assumed to be fixed to values learned at test time, and Eqs.~(\ref{eq:update_a}) and (\ref{eq:update_a2}) are not updated for test images.}

\noindent\textbf{Free image annotation: } This is to describe an image using a list of nouns and adjectives corresponding to objects and their associated attributes. To infer the objects  present in  image $i$, the first $K_o$ latent factors of the inferred $\bm{\pi}^{(i)}$ are thresholded or ranked to obtain a list of objects. This is followed by locating them via searching for the superpixels $j^*$ maximising $\mathbf{Z}^{(i)}_{jk}$, then thresholding or ranking the $K_a$ attribute latent factors in $\mathbf{Z}^{(i)}_{j^*k}$ to describe them. %\textcolor{blue}{Note that $\mathbf{Z}^{(i)}_{jk}$ is inferred with the learned $\mathbf{A}$. Thus we do not need to update Eq.~\ref{eq:update_a},\ref{eq:update_a2} for testing image.}
This corresponds to a ``\emph{describe this image}'' task.

%, and then describing them by the second $K_a$ latent factors of each $\mathbf{Z}^{(i)}_j$
%\textbf{Conventional Annotation: } Conventional (without association) image-level annotation for each image $i$ can be performed using the first $K_o$ latent factors of the inferred $\bm{\pi}^{(i)}$, and for attributes by the second $K_a$ latent factors of $\bm{\pi}^{(i)}$. 
\noindent\textbf{Annotation given object names:} This is a more constrained variant of the free annotation task above. Given a named (but not located) object $k$, its associated attributes can be estimated by first finding the location as $j^*=\underset{j}{\arg\max} \  \bm{Z}^{(i)}_{jk}$, then the associated attributes  by $\bm{Z}^{(i)}_{j^*k}$ for $K_o<k \leq K_o+K_a$. This corresponds to a ``\emph{describe this (named) object in an image}'' task.

\noindent\textbf{Object+attribute query: } Images can be queried for a specified object-attribute conjunction $<k_o, k_a>$ by searching for $i^*,j^*=\underset{j}{\arg\max} \  \bm{Z}^{(i)}_{jk_o} \cdot \bm{Z}^{(i)}_{jk_a}$. %\textcolor{blue}{Similar as free annotation, $(\bm{\phi}_k, \bm{\Phi}_k)$ is fixed to the learned A. }
This corresponds to a ``\emph{find images with a particular \textbf{kind of} object}'' task.

\noindent\textbf{Semantic segmentation:} In this application, we aim to label each superpixel $j$ with one of $K_o$ learned object factors. The label of superpixel $j$ can be obtained by searching  $k^*=\underset{k}{\arg\max} \ \bm{Z}^{(i)}_{jk}$, where $k \in K_o$. Although the annotation search space is solely objects, inference of the additional $k > K_o$ factors (including unannotated background or attribute annotation) can help detect objects $k\in K_o$ via disambiguation. Note that unlike most weakly supervised semantic segmentation methods \cite{xu_cvpr2014,Yang_cvpr2013}, our model can operate \emph{without} access to the whole test set. But it can also operate in a transductive setting as those existing methods. Under this setting, the appearance distribution $\mathcal{N}(\bm{A}_{k\cdot}; \bm{\phi}_k, \bm{\Phi}_k)$ will be further updated by Eqs.~(\ref{eq:update_a}) and (\ref{eq:update_a2}) based on the test images. The image-level labels of test data is assigned by the inferred factors of our model or alternatively by an image classifier (see Sec.~\ref{sec:labelMe}).

\section{Experiments}
\label{sec:exp}
Extensive experiments are carried out to demonstrate the effectiveness of our model on three real-world applications: automatic image annotation (see Sec.~\ref{sec:annotation}), object-attribute query (see Sec.~\ref{sec:query}) and semantic segmentation (see Sec.~\ref{sec:segment}).

\begin{figure*}[t]

\begin{minipage}{0.6\textwidth}
\raggedright
  \includegraphics[width=1.0\linewidth]{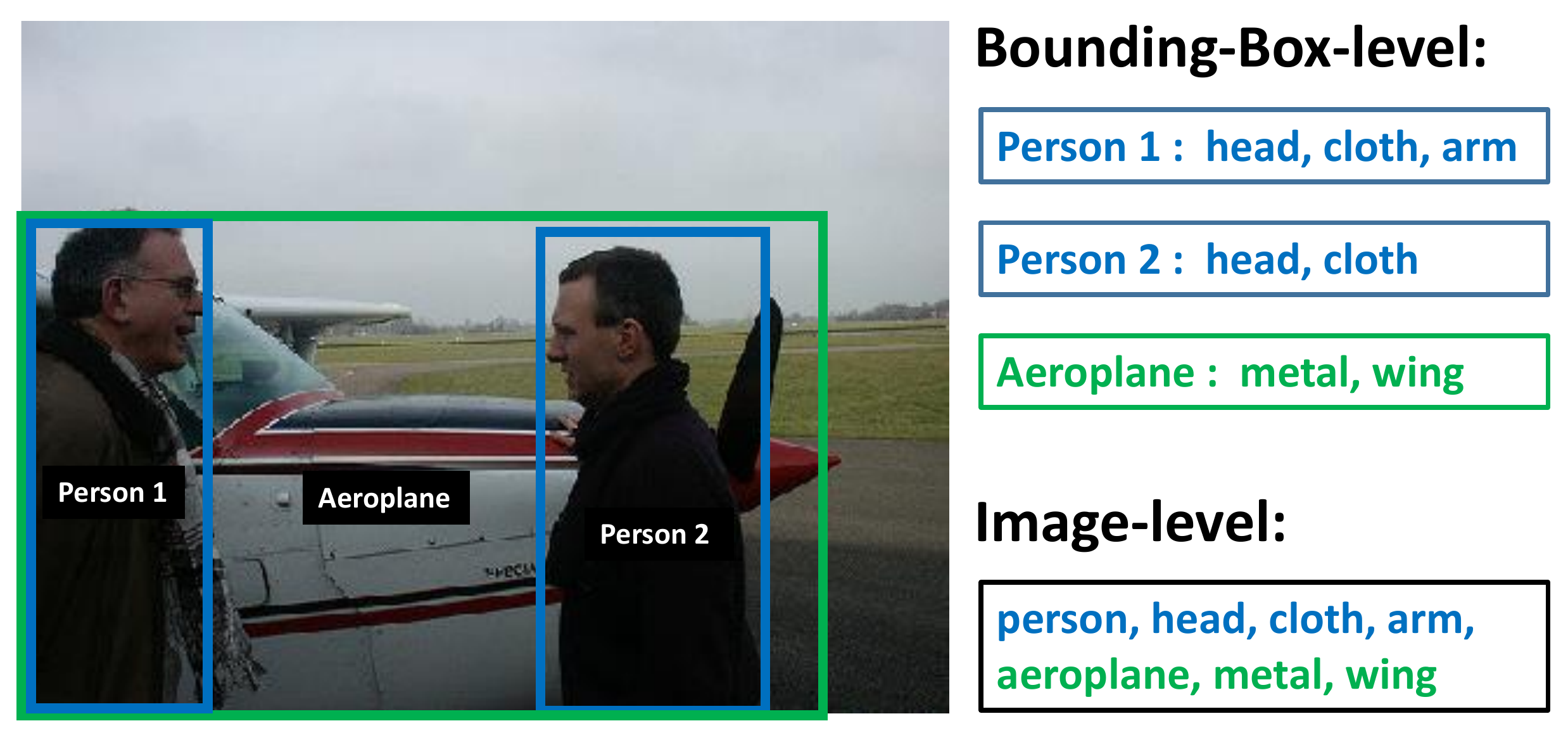}
  \captionof{figure}{Strong bounding-box-level annotation and weak image-level annotations for aPascal are used for learning strongly supervised models and weakly supervised models respectively.}
  \label{fig:dataset_annotat}
\end{minipage}%
\hspace{0.25cm}
\begin{minipage}{.34\textwidth}
\raggedleft
  \includegraphics[width=1.0\linewidth]{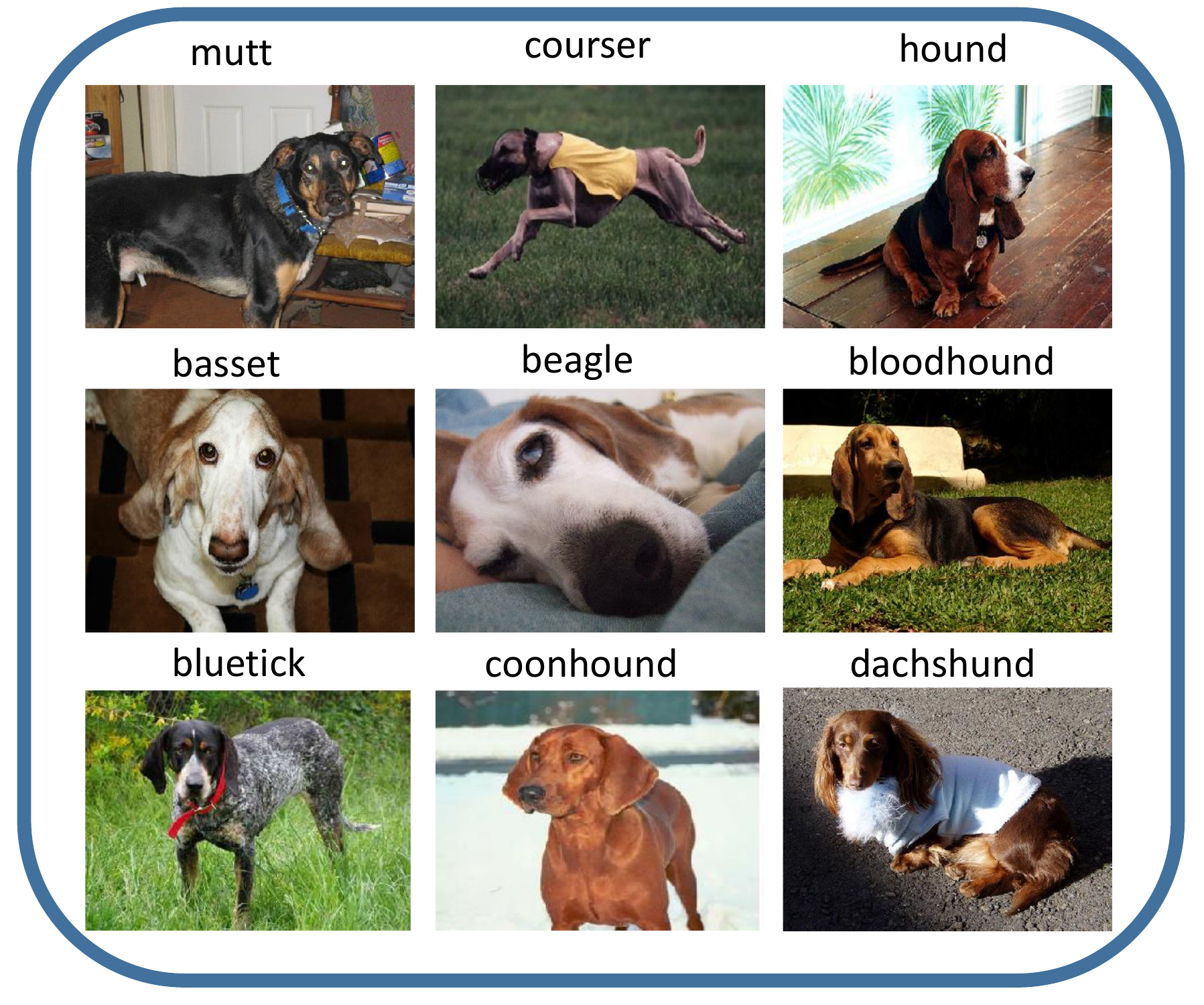}
%  \captionof{figure}{ImageNet Attribute dataset contains 43 subordinate classes of dog which are converted into a single  entry-level class `dog'.}
  \captionof{figure}{43 subordinate classes of dog are converted into a single  entry-level class `dog'.}
   \label{fig:dataset_entry}
\end{minipage}
\end{figure*}

\subsection{Image annotation and query }
\label{sec:image_anno}

\subsubsection{Datasets and settings}
\label{sec:annotation setting}

For the automatic image annotation and query tasks, various object and attribute datasets are available such as aPascal \cite{Farhadi09CVPR}, ImageNet \cite{RussakovskyECCV10}, SUN \cite{Patterson_2012_CVPR} and AwA \cite{lampert13AwAPAMI}. We choose aPascal because it has multiple objects per image; and ImageNet because attributes are shared widely across categories. %For semantic segmentation task, LabelMe Outdoor \cite{Liu2011} has been widely used in recent studies. We also evaluate the aPascal segmentation dataset \cite{Zheng_cvpr2014} as both attribute and pixel-level annotations are available.

\noindent\textbf{aPascal:}\quad This dataset \cite{Farhadi09CVPR} is an attribute labelled version of PASCAL VOC 2008. There are 4340 images of 20 object categories. Each object is annotated with a list of 64 attributes that describe them by shape (e.g., isBoxy), parts (e.g., hasHead) and material (e.g., isFurry). In the original aPascal, attributes are strongly labelled for 12695 object bounding boxes, i.e.~the object-attribute association are given. To test our weakly supervised approach, we merge the object-level category annotations and attribute annotations into a single annotation vector of length 84 for the entire image. This image-level annotation is much weaker than the original bounding-box-level annotation, as shown in Fig.~\ref{fig:dataset_annotat}. In all experiments, we use the same train/test splits provided by \cite{Farhadi09CVPR}. 

\noindent\textbf{ImageNet Attribute:} This dataset \cite{RussakovskyECCV10} contains 9600 images from 384 ImageNet synsets/categories. To study WSL, we ignore the provided bounding box annotation. Attributes for each bounding box are labelled as 1 (presence), \hspace{0.2cm} -1\\ (absence) or 0 (ambiguous). We use the same 20 of 25 attributes as \cite{RussakovskyECCV10} and consider 1 and 0 as positive examples. Many of the 384 categories are subordinate categories, e.g.~dog breeds. However, distinguishing  fine-grained subordinate categories is beyond the scope of this study. That is,  we are interested in finding a `black-dog' or `white-car', rather than `black-labrador' or `white-ford-focus'. We thus convert the 384 ImageNet categories to 172 entry-level categories using \cite{Ordonez_2013} (see Fig.~\ref{fig:dataset_entry}). We evenly split each class to create the training and testing sets.

 We compare our WS-MRF-SIBP to two strongly supervised models and four weakly supervised alternatives:

\noindent \textit{\textbf{Strongly supervised models}}: A strongly supervised model uses bounding-box-level annotation. Two variants are considered for the two datasets respectively. \textbf{DPM+s-SVM}: for aPascal, both object detector and attribute classifier are trained from fully supervised data (i.e.~Bounding-Box-level annotation in Fig.~\ref{fig:dataset_annotat}). Specifically, we use the  20 pre-trained DPM detectors from \cite{lsvm_pami} and 64 attribute classifiers from \cite{Farhadi09CVPR}. \textbf{GT+s-SVM}: for ImageNet attributes, there is not enough data to learn 172 strong  DPM detectors as in aPascal.  So we use the ground truth bounding box instead assuming we have perfect object detectors, giving a significant advantage to this strongly supervised model. We train attribute classifiers using our features (Sec.~\ref{sec:representation}) and liblinear SVM \cite{REF08a}. These strongly supervised models are similar in spirit to the models used in \cite{kulkarni2011descriptions,wang2013wslAttrLoc,wang2009attrib_class_sal} and can be considered to provide an upper bound for the performance of the weakly supervised models.

\noindent \textit{\textbf{Weakly supervised models}}: \noindent\textbf{w-SVM} \cite{Farhadi09CVPR,RussakovskyECCV10}:~~In this weakly-supervised baseline, both object detectors and attribute classifiers are trained on the weak image-level labels as for our model (see Fig.~\ref{fig:dataset_annotat}). For aPascal, we train object and attribute classifiers using the feature extraction and model training codes (which is also based on \cite{REF08a}) provided by the authors of \cite{Farhadi09CVPR}. For ImageNet, our features are used, without segmentation.
\textbf{MIML}:~~This is the multi-instance multi-label (MIML) learning method in \cite{Zhou20122291}. Our model can also be seen as a MIML method with each image a bag and each superpixel an instance. The MIML model  provides a mechanism to use the same superpixel based representation for images as our model, thus providing the object/attribute localisation capability as our model does. 
 \textbf{w-LDA}:~~Weakly-supervised Latent Dirichlet Allocation (LDA) approaches \cite{Rasiwasia_2013,Shi_2013_ICCV} have been used for object localisation. We implement a  generalisation of LDA \cite{blei2003lda,Shi_2013_ICCV} that accepts continuous feature vectors (instead of bag-of-words). Like MIML this method can also accept superpixel based representation, but w-LDA is more related to our WS-SIBP than MIML since it is also a generative model. %But as explained in Sec.~\ref{sec:related work},  LDA based models suffer from a pre-allocated number of topics, and the implicit assumption of a single label per segment. 
\textbf{WSDC} \cite{Yang_cvpr2013}: Weakly supervised dual clustering is a recently proposed method for semantic segmentation that estimates pixel-level annotation given only image-level labels. This semantic segmentation method can be re-purposed to our image annotation setting by considering the same input (superpixel representation + image-level label) followed by the same method as in our framework to first infer superpixel level labels and then aggregate them to compute image-level annotations (see Sec.~\ref{sec:postproc}).

\begin{figure*}[t]
\centering
%\fbox{\includegraphics[height=3.5cm]{ori_A.pdf}}
\includegraphics[width=1.0\linewidth]{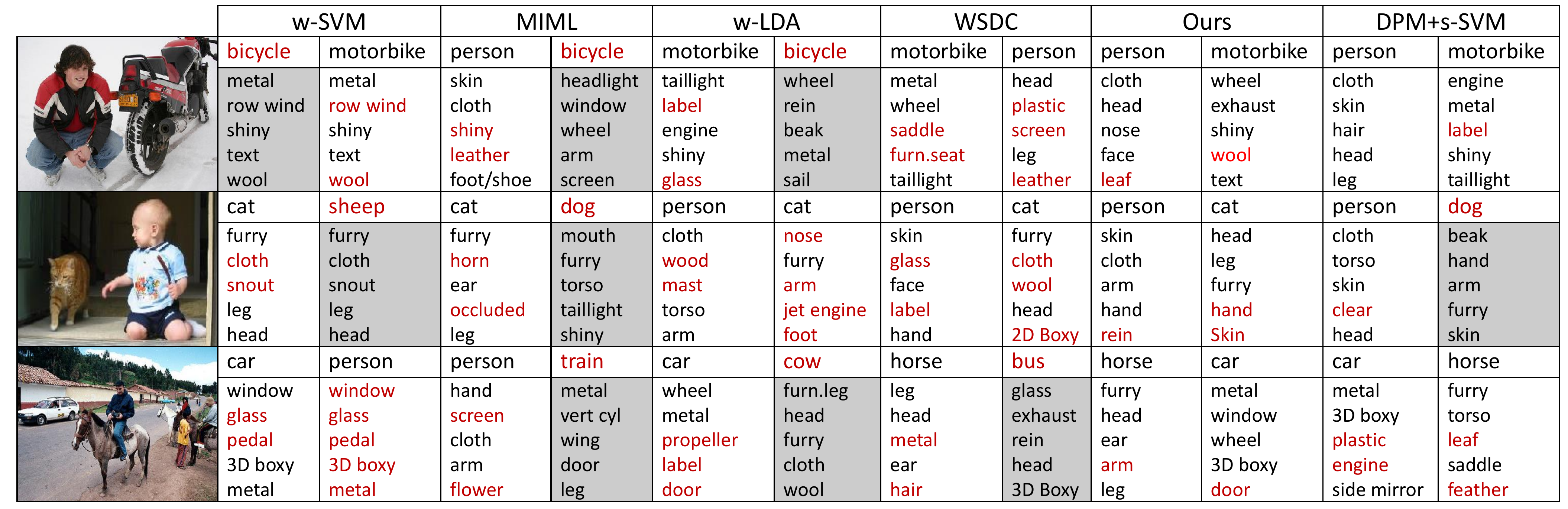}
%\label{1a}
\vskip -0.2cm
\caption{Qualitative results on free annotation. False positives are shown in red. If the object prediction is wrong, the corresponding attribute box is shaded.}
\label{fig:exp1_real_vis}
\end{figure*}

\subsubsection{Automatic Image annotation}
\label{sec:annotation}

An image description (annotation) can be automatically generated by predicting objects and their associated attributes.  To comprehensively cover all aspects of performance of our method and competitors, we perform three annotation tasks with different constraints on test images: (1) \textit{free annotation}, where no constraint is given to a test image, (2) \textit{annotation given object names}, where named but not located objects are given, and  (3) \textit{annotation given locations}, where object locations are given in the form of bounding boxes, and the attributes are predicted.

\begin{table}[h]
%\footnotesize
%\small
%\begin{tabular*}{5cm}{@{\extracolsep{\fill}}lllr}
\setlength{\tabcolsep}{0.25em}
\centering
{%\footnotesize
%\begin{tabular}{l || l | l | l ||  l  || l  }
\resizebox{1.0\columnwidth}{!}{
\begin{tabular}{l | l | l | l   | l | l | l   }

\hline

 & \multicolumn{3}{c|}{aPascal \cite{Farhadi09CVPR}} & \multicolumn{3}{c}{ImageNet \cite{RussakovskyECCV10}}  \\

\cline{2-7}
&  AP@2  &  AP@5  &  AP@8   &  AP@2  &  AP@5  &  AP@8  \\

\cline{1-7}
w-SVM \cite{Farhadi09CVPR} &  24.8 & 21.2 &  20.3  & 46.3 & 41.1 &  37.5 \\ 
%\hline
MIML \cite{Zhou20122291} &   28.7 & 22.4 & 21.0 &  46.6 & 43.2 & 38.3 \\ 
%\hline
w-LDA \cite{Shi_2013_ICCV}  &    30.7  & 24.0 & 21.5  & 48.4  & 43.1 & 38.4   \\ 
WSDC \cite{Yang_cvpr2013}  & 29.8 & 25.1 & 21.3 & 48.0 & 42.7 & 36.5  \\ 

\hline
%Our \cite{Shi_2014_ECCV}  &  38.6  & 28.9 & 24.1  &  58.5  & 51.8 & 47.4  \\ 

Ours   & \textit{40.1}  &  \textit{29.7}  & \textit{\textbf{25.0}}   & \textit{60.7 } & \textit{54.2}    & \textit{50.0}   \\ 
\hline
D/G+s-SVM   & \textbf{40.6} &  \textbf{30.3}  & 23.8  & \textbf{65.9} &  \textbf{60.7}  & \textbf{53.2}   \\

\hline
\end{tabular}}
}
\caption{Free annotation performance (AP@t) evaluated on $t$ attributes per object.}
\label{tab:exp1_real}
\end{table}

\noindent \textbf{Free annotation:} For  WS-MRF-SIBP, w-LDA and MIML the procedure in Sec.~\ref{sec:postproc} is used to  detect objects and then describe them using the top $t$ attributes. %A similiar precedure is adopted for w-LDA and MIML. 
For the strongly supervised model on aPascal (DPM+s-SVM), we use DPM object detectors to find the most confident objects and their bounding boxes in each test image. Then we use the 64 attribute classifiers to predict top $t$ attributes in each bounding box. In contrast,
 %to the other four models that localise objects and associate their attributes, 
 w-SVM trains attributes and objects independently, and cannot associate objects and attributes. We thus use it to predict only one attribute vector per image regardless of which object label it predicts. 

Since there are a variable number of objects per image in aPascal, quantitatively evaluating free annotation is not straightforward. 
%\textcolor{cyan}{In particular, we do not want to use the ground truth object number as the number of objects generated by the compared  models, because this number is not available during testing in a practical application scenario.} 
Therefore, we evaluate only the most confident object and its associated top $t$ attributes in each image, although more could be described. For ImageNet, there is only one object per image. %, so this problem goes away.  
We follow \cite{Feng_2013_ICCV,tag_2013} in evaluating annotation performance by average precision (AP@t), given varying numbers ($t$) of predicted attributes per object. If the predicted object is wrong, all associated attributes are considered wrong.

Table~\ref{tab:exp1_real} reports the free annotation performance of the compared models. We have the following observations: (1) Our WS-MRF-SIBP, despite learned from weak image-level annotation, yields comparable performance to the strongly supervised model (DPM/GT+s-SVM). The gap is particularly small for the more challenging aPascal dataset, whist for ImageNet, the gap is bigger as the strongly supervised GT+s-SVM has an unfair advantage  by using the ground truth bounding boxes during testing. (2) WS-MRF-SIBP consistently outperforms the four weakly supervised alternatives. The margin is especially large for $t=2$ attributes per object, which is closest to the true number of attributes per object. For bigger $t$, all models must generate some irrelevant attributes thus narrowing the gaps.  (3) As expected, the w-SVM model obtains the weakest results, confirming that the ability to locate objects is important for modelling object-attribute association. (4) Compared to the two generative models (ours and w-LDA), MIML has worse performance because a generative model is more capable of utilising weak labels \cite{Shi_2013_ICCV}. The other discriminative model WSDC fares better than MIML due to its ability to exploit superpixel appearance similarity to disambiguate  image-level labels, but it is still inferior to our model. (5) Between the two generative models, the advantage of our framework over w-LDA is clear; due to the ability of IBP to explain each superpixel with multiple non-competing factors\footnote{Training two independent w-LDA models for objects and attributes respectively is not a solution: the problem would re-occur for multiple competing attributes.}.

%\textcolor{red}{One might argue that training two independent w-LDA models may solve the issue of competing between object and attribute, while the same problem would re-occur within the attribute model for multiple attribute competing.}
% factors for explaining the superpixel appearance, and allowing multiple factors co-existing rather than competing with each other.  

%\begin{figure}[t]
%\centering
%%\fbox{\includegraphics[height=3.5cm]{ori_A.pdf}}
%\includegraphics[width=\linewidth]{fig/fig5_exp2.pdf}
%%\label{1a}
%\vskip -0.2cm
%\caption{Qualitative results on free annotation. False positives are shown in red. If the object prediction is wrong, the corresponding attribute box is shaded. }
%\label{fig:exp1_real_vis}
%\end{figure}

Fig.~\ref{fig:exp1_real_vis} shows some qualitative results on aPascal via the two most confident objects and their associated attributes.
%Qualitative results on aPascal are shown in Fig.~\ref{fig:exp1_real_vis}: the two most confident objects and their associated attributes are given. 
This is a challenging dataset -- even the strongly supervised DPM+s-SVM makes mistakes for both attribute and object prediction. Compared to the other weakly supervised models, WS-MRF-SIBP has more accurate predictions --  it jointly and non-competitively models objects and their attributes so object detection benefits from attribute detection and vice versa.  %This benefit diminishes for w-LDA because object topics  compete with attribute topics for the same patch. 
The competitors are also more likely to mismatch attributes with objects, e.g.~MIML detects a shiny person rather than the correct shiny motorbike. 

\begin{figure*}[t]
\centering
%\fbox{\includegraphics[height=3.5cm]{ori_A.pdf}}
\includegraphics[width=1.0\linewidth]{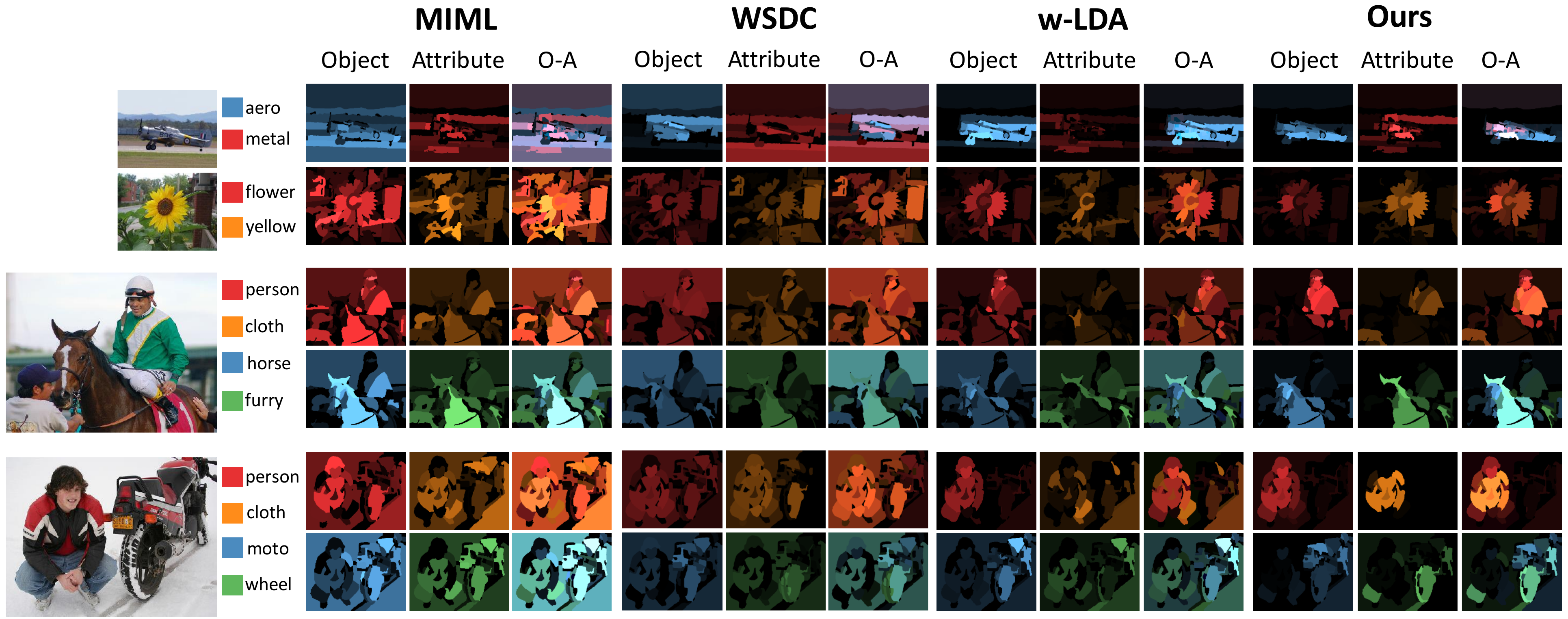}
%\label{1a}
\caption{Illustrating the inferred superpixel annotation. Object and attributes are coloured, and multi-label annotation blends colours. Each of the bottom two groups has two rows corresponding to the two most confident objects detected.}
\label{fig:obj_att_topic}
\end{figure*}

To gain  insight into what has been learned by our model and why it is better than the weakly supervised alternatives, Fig.~\ref{fig:obj_att_topic} visualises the attribute and object factors learned by the competing models which  use  superpixels as  input. It is evident that without explicit  background modelling, MIML suffers  by trying to explain the background superpixel using the weak labels. In contrast, both w-LDA and WS-SIBP have good segmentation of  foreground objects, showing that both the learned foreground and background topics are meaningful. However, for w-LDA, since object and attributes topics compete for the same superpixel, each superpixel is dominated by either an object  or attribute topic. In contrast, the object factors and attribute factors co-exist happily in WS-SIBP as they should do, e.g.~most person superpixels have the clothing attribute as well. 

\noindent \textbf{Annotation given object names (GN):} In this experiment, we assume that object labels are given and we aim to describe each object by attributes, corresponding to tasks such as: ``\emph{Describe the car in this image}". 
For the strongly supervised model on aPascal, we use the object's DPM detector to find the most confident bounding box. Then we predict attributes for that box. Here,  annotation accuracy is the same as attribute  accuracy, so the performance of different models is evaluated following \cite{Zhang_2013_ICCV} by mean average precision (mAP) under the precision-recall curve. Note that for aPascal,  w-SVM reports the same list of attributes for all co-existing objects, without being able to localise and distinguish them. Its result is thus not meaningful and is excluded. The same set of conclusions can be drawn from Table~\ref{tab:att_pre_apascal} as in the free annotation task: our WS-MRF-SIBP is on par with supervised models and outperforms  weakly supervised ones.

\begin{table}[h]
\setlength{\tabcolsep}{0.3em}
\small
\centering
%{\footnotesize
%\begin{tabular}{l|l || l | l | l || l || l  }
%\scalebox{0.95}{
\begin{tabular}{l|l | l  l  l l | l  | l  }
\hline
&&  w-SVM  & MIML & w-LDA & WSDC &  Ours   & SS  \\
%\hline
\hline \parbox[t]{3mm}{\multirow{2}{*}{\rotatebox[origin=c]{90}{GN}}} &aPascal   & -- & 32.1 & 35.5  & 36.3 &\textit{39.3}& \textbf{41.8} \\ 
% 35.8$^*$
%\cline{2-7}
%aPascal-1000&   &   & &   &   \\ 
%\hline
&ImageNet&  32.4  & 33.5  & 39.6 & 44.2 &  \textit{52.8}& \textbf{56.8}\\ 
%\hline
\hline
%\textit{Siva \etal} \cite{confeccvSivaRX12}  & 27.1\%   &   28.0 \% \\\hline
\parbox[t]{3mm}{\multirow{2}{*}{\rotatebox[origin=c]{90}{GL}}} &aPascal   &  33.2& 35.1 & 35.8 & 38.4 & \textbf{\textit{43.6}} & 42.1   \\ 
%\cline{2-7}
&ImageNet & 37.7  &  39.1  &  46.8  & 48.2 &  \textit{53.9} &  \textbf{56.8}  \\ 
\hline
\end{tabular}
%}
%}
\caption{Results on annotation given object names (GN) or locations (GL). SS stands for Strongly Supervised. }

\label{tab:att_pre_apascal}
\end{table}

\begin{figure*}[t]
\centering
\subfigure[O-A, aPascal]{
  \centering
  \includegraphics[height=.253\linewidth]{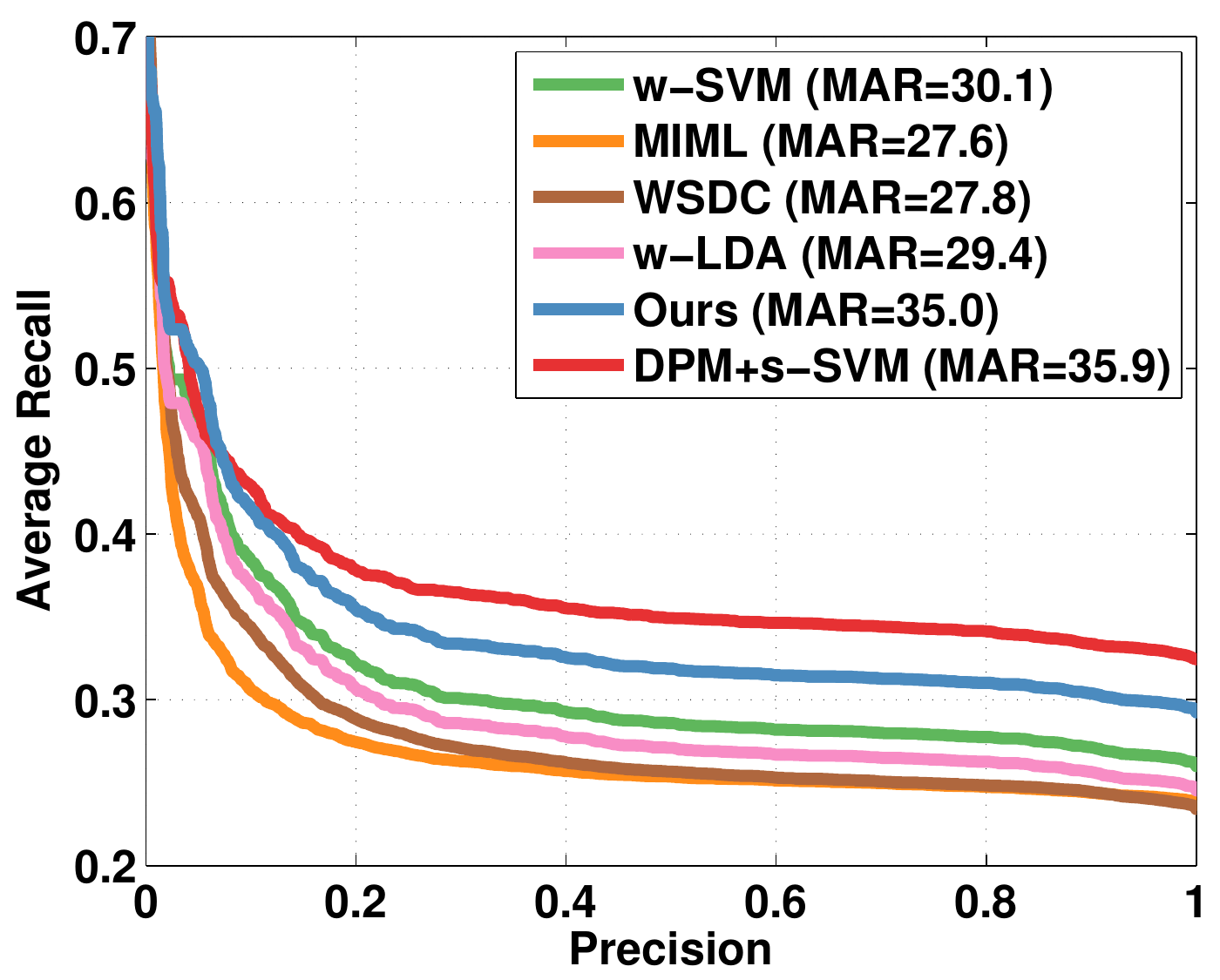}
  %\caption{aPascal}
  \label{fig:sub1}
}%
\subfigure[O-A, ImageNet]{
  \centering
  \includegraphics[height=.253\linewidth]{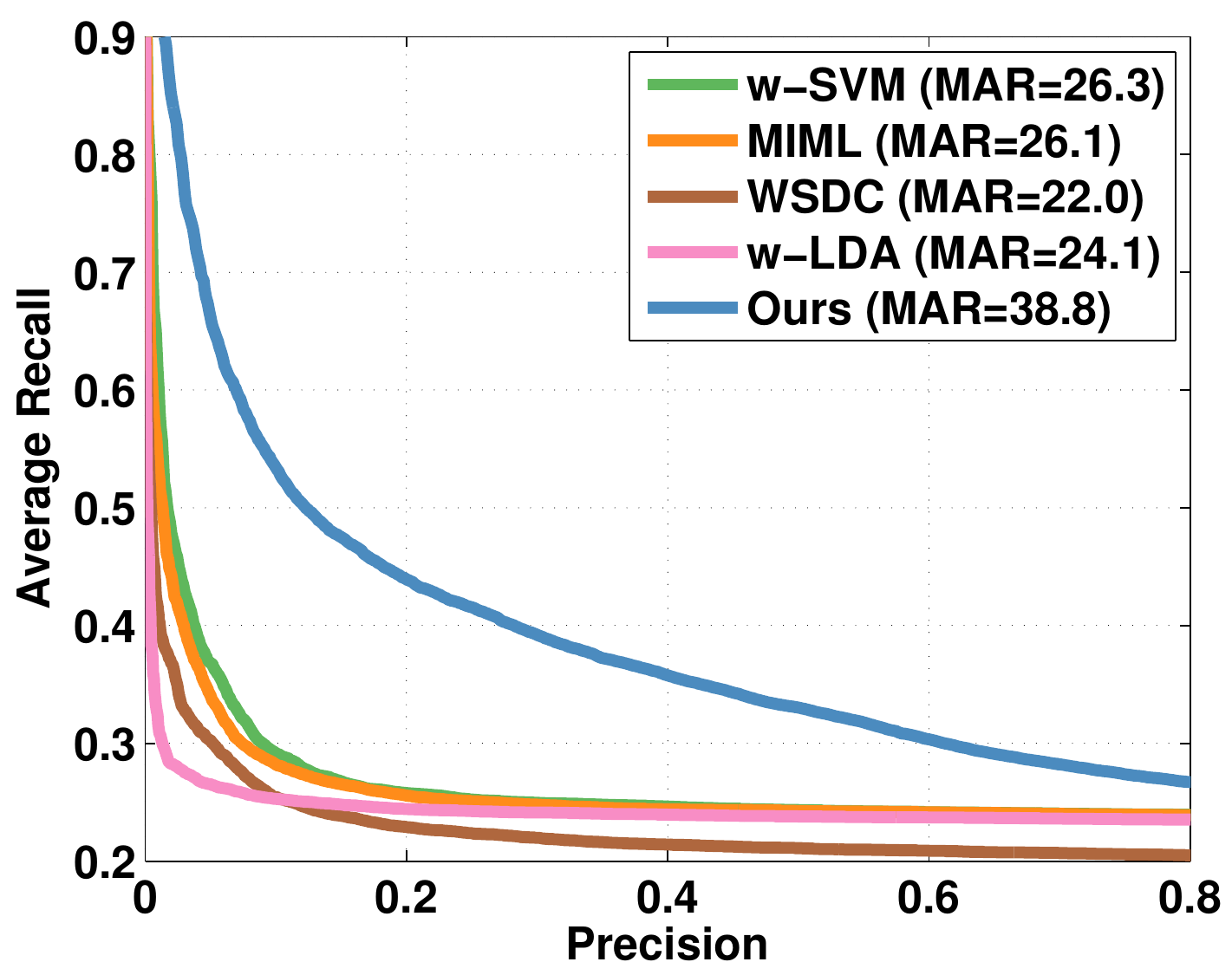}
  %\caption{aPascal}
  \label{fig:sub1}
}%
\subfigure[O-A-A, ImageNet]{
  \centering
  \includegraphics[height=.253\linewidth]{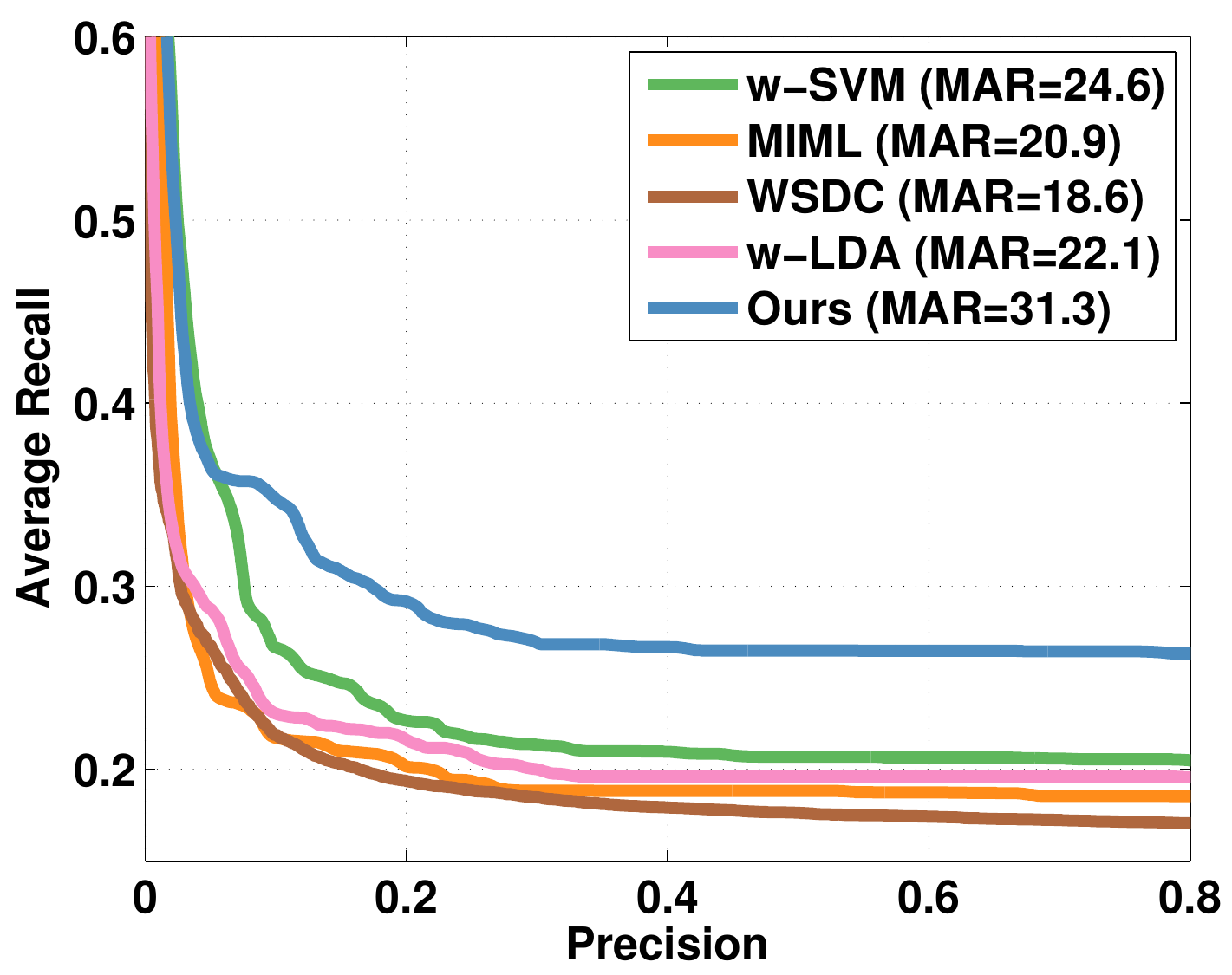}
  %\caption{aPascal}
  \label{fig:sub1}
}%
\caption{Object-attribute query results as precision-average recall curve.}
%\caption{Object-attribute query results as precision-average recall curve (MAR values are shown in brackets).}
\label{fig:test}
\end{figure*}

\noindent \textbf{Given object location (GL):} If we further know the bounding box of an object in a test image, we can simply predict attributes inside each bounding box. This becomes  the conventional attribute prediction task \cite{Farhadi09CVPR,RussakovskyECCV10} for describing an object. Table~\ref{tab:att_pre_apascal} shows the results, where similar observations can be made as in the other two tasks above. Note that in this case  the strongly supervised model  is the method used in \cite{Farhadi09CVPR}. The mAP obtained using our weakly supervised model is even higher than the strongly supervised model (though our area-under-ROC-curve value of 81.5 is slightly lower than the 83.4 reported in \cite{Farhadi09CVPR}).

\subsubsection{Object-attribute query}
%\subsection{Object-attribute query}
\label{sec:query}

In this task object-attribute association is used for image retrieval.  Following work on multi-attribute queries \cite{Rastegari_CVPR13}, we use mean average recall over all precisions (MAR) as the evaluation metric. Note that unlike \cite{Rastegari_CVPR13} which requires each queried \emph{combination} to have enough (100) training examples to train conjunction classifiers, our method can query novel never-previously-seen combinations. Three experiments are conducted. We generate 300 random object-attribute combinations for aPascal and ImageNet respectively and 300 object-attribute-attribute queries for ImageNet. For the strongly supervised model, we normalise and multiply  object detector with attribute classifier scores. No object detector is trained for ImageNet so no result is reported there. For w-SVM, we use  \cite{multiattrs_cvpr2012} to calibrate the SVM scores for objects and attributes as in \cite{Rastegari_CVPR13}. For the three WS models, the procedure in Sec.~\ref{sec:postproc} is used to compute the retrieval ranking.

%\vspace{10cm}

\setlength{\belowcaptionskip}{-13pt}

\begin{figure*}[t]
\centering
%\fbox{\includegraphics[height=3.5cm]{ori_A.pdf}}
\includegraphics[width=0.8\linewidth]{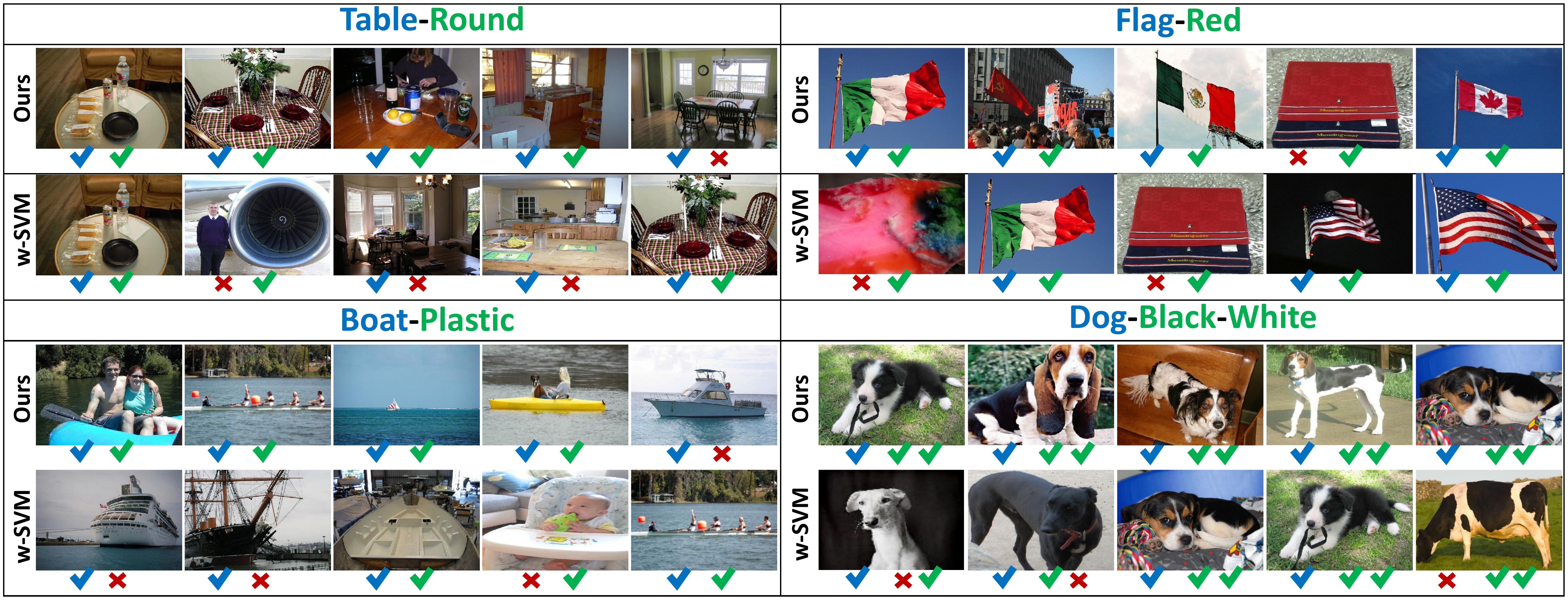}
%\label{1a}
\caption{Object-attribute query: qualitative comparison. \textcolor{black}{Given a query, such as ``Table+Round'', we list the top-5 results predicted by our method and w-SVM.}}
\label{fig:object_att_ann}
\end{figure*}

Quantitative results are shown in Fig.~\ref{fig:test} and some qualitative examples  in Fig.~\ref{fig:object_att_ann}. Our approach has very similar MAR values to the strongly supervised DPM+s-SVM, while outperforming all the other models. 
%Note that for this task, we are able to calibrate w-SVM scores using  \cite{multiattrs_cvpr2012}, explaining why w-SVM is now better than MIML and w-LDA.
w-SVM calibration  \cite{multiattrs_cvpr2012} helps it outperform MIML and w-LDA.
%Note that for this task, localising objects and attributes are less critical compared to the tasks of object-attribute association based image annotation. w-SVM benefits the most from this less stringent condition.  This explains why the result of w-SVM is better than that of MIML and w-LDA now.  
However, the lack of object-attribute association and background modelling still causes problems for w-SVM. This is illustrated in the `dog-black-white' example shown in Fig.~\ref{fig:object_att_ann} where a white background caused an image with a black dog retrieved at Rank 2 by w-SVM.

\subsection{Semantic Segmentation}
\label{sec:segment}

\subsubsection{Datasets and settings}

We evaluate  semantic segmentation performance on the aPascal Segmentation  (Sec.~\ref{sec:apascal_seg}) and LabelMe Outdoor datasets (Sec.~\ref{sec:labelMe}) under the weakly supervised setting.

\noindent\textbf{aPascal Segmentation:}  This dataset \cite{Zheng_cvpr2014} is a subset of PASCAL VOC 2008 \cite{Farhadi09CVPR} where both pixel-level segmentation and attributes annotation are available. It contains 639 images from 20 classes. The 64 attribute annotation for each image is the same as the  aPascal dataset used in the annotation experiments. We use the  training (326 images) and testing (313 images) split provided by \cite{Zheng_cvpr2014}.

\noindent\textbf{LabelMe Outdoor Segmentation:} Also known as  SIFT Flow  \cite{Liu2011}, this widely used dataset contains 2688 images densely labelled with 33 object classes at pixel-level using the LabelMe online annotation tool.  Every pixel in each image is assigned a label meaning that background `stuff' such as sky, sea, street  are also labelled as objects. Most images contain outdoor scenes. We use the standard training (2488 images) and testing (200 images) split provided in \cite{Liu2011}. Note that no attribute labels are available for this dataset.

\noindent\textbf{Evaluation metrics:} The evaluation metrics used for semantic segmentation are often dataset dependent. \textcolor{black}{Past works \cite{Tighe_IJCV2015, Stephen_eccv2014} on the LabelMe dataset typically report results in both total per-pixel accuracy (defined as the number of correctly labelled pixels over the total number of pixels), and per-class accuracy (defined as the number of correctly labelled pixels for a class over the number of ground truth pixels of this class and then averaged over all object classes).} Both metrics are necessary because for any model, some parameters can typically be tuned so that one metric is favoured  at the expense of the other. The  VOC images have very different characteristics compared with LabelMe. In particular, the images often contain large portions of unannotated background (stuff) and the 20 objects of interest are relatively small. The intersection-over-union (IOU) score is thus typically  used for semantic segmentation performance evaluation on the Pascal VOC dataset \cite{Zheng_cvpr2014,Yang_cvpr2013,Dong_eccv2014} and adopted here on aPascal. 

\subsubsection{Results on aPascal }
\label{sec:apascal_seg}

To our knowledge, no previous work models objects and attributes jointly for semantic segmentation in the weakly supervised setting. We therefore apply the  weakly supervised (object only) segmentation method WSDC \cite{Yang_cvpr2013} as an alternative\footnote{We do not have access to the codes of other weakly supervised  segmentation methods and implementing them is non-trivial.} (see Sec.~\ref{sec:annotation setting}). With our WS-MRF-SIBP, the association of objects and available attributes can be leveraged to improve  performance. We explore three different  sets of attribute annotations: (1) 8 attributes:  material  attributes used by \cite{Zheng_cvpr2014} for  object-attribute segmentation. (2) 64 attributes: the original attributes provided by \cite{Farhadi09CVPR}. (3) 74 attributes: we add 10 more color attributes based on aPascal sentence descriptions \cite{kulkarni2011descriptions,Rashtchian_2010}. 

\begin{table}[H]
%\scriptsize
%\setlength{\tabcolsep}{0.15em}
%\setlength\extrarowheight{1pt}
\small
\centering
%{\footnotesize
%\begin{tabular}{l || l | l | l ||  l  || l  }
%\scalebox{0.95}{
\begin{tabular}{  M{0.3cm} |  l | c   }
\hline
\multicolumn{2}{c |}{Method} &  Avg. IOU (\%)   \\
\hline

\parbox{3mm}{\multirow{2}{*}[-4pt]{\rotatebox{90}{Fully}}} & Zheng et al \cite{Zheng_cvpr2014}  & \textbf{37.1} \\
\cline{2-3}
& Kr\"{a}henb\"{u}hl  et al \cite{NIPS2011_Philipp}  & 36.9 \\
\hline
\hline
\parbox{3mm}{\multirow{5}{*}[-4pt]{\rotatebox{90}{Weakly}}} & WSDC \cite{Yang_cvpr2013}   & 18.2\\
\cline{2-3}
& Ours\_0attribute & \textit{23.6}\\
\cline{2-3}
& Ours\_8attributes  & \textit{27.3} \\
\cline{2-3}
& Ours\_64attributes  & \textit{28.9} \\
\cline{2-3}
& Ours\_74attributes  & \textbf{\textit{29.4}} \\
\hline
\end{tabular}%}
%}
\caption{Quantitative semantic segmentation comparison versus state-of-the-art on the aPascal dataset. }
\label{tab:apascal_seg}
\end{table}

Our model is compared with one weakly supervised \cite{Yang_cvpr2013} and two fully-supervised alternatives \cite{Zheng_cvpr2014,NIPS2011_Philipp} in Table~\ref{tab:apascal_seg}. It can be observed that our method outperforms the alternative weakly-supervised model WSDC \cite{Yang_cvpr2013}, even without attribute annotation (Ours\_0attribute). Moreover, its performance gradually improves as more attribute annotation becomes available, and eventually its performance with 74 attributes is not far off from the two fully-supervised models. % Note that part of the reason for our good performance in the low, or no attribute, case is that  the discovered latent factors are available to model unannotated attributes. Thus attributes detectable from data can still  be exploited to some extent to help object disambiguation, even when they are not annotated.

Some qualitative results are shown in Fig.~\ref{fig:apascal}. In each example, we show the color-coded segmentation output of Ours\_0attribute, Ours\_64attribute, Kr\"{a}henb\"{u}hl et al \cite{NIPS2011_Philipp} and ground truth segmentation. Coloured regions are identified as the same foreground objects. Background clutter with similar appearance to foreground object can confuse Ours\_0attribute as shown in the horse example (the third row of Fig.~\ref{fig:apascal}). However, by exploiting the additional (weak) attribute annotation, the segmentation performance is greatly improved (Our\_64attribute) through disambiguation, and by capturing object-attribute co-occurrence. 

%1) Most of previous works are transductive based model. They can not deal with single image. Our model
%can do it without transductive setting.
%2) Our model can learn some latent attributes which co-occur with some objects. e.g. green color is strongly
%correlated with grass.
%3) If the attribute annotation is available, our model can utilize it to further improve the segmentation
%results.

\begin{figure}[t]
\centering
%\fbox{\includegraphics[height=3.5cm]{ori_A.pdf}}
\includegraphics[width=1\linewidth]{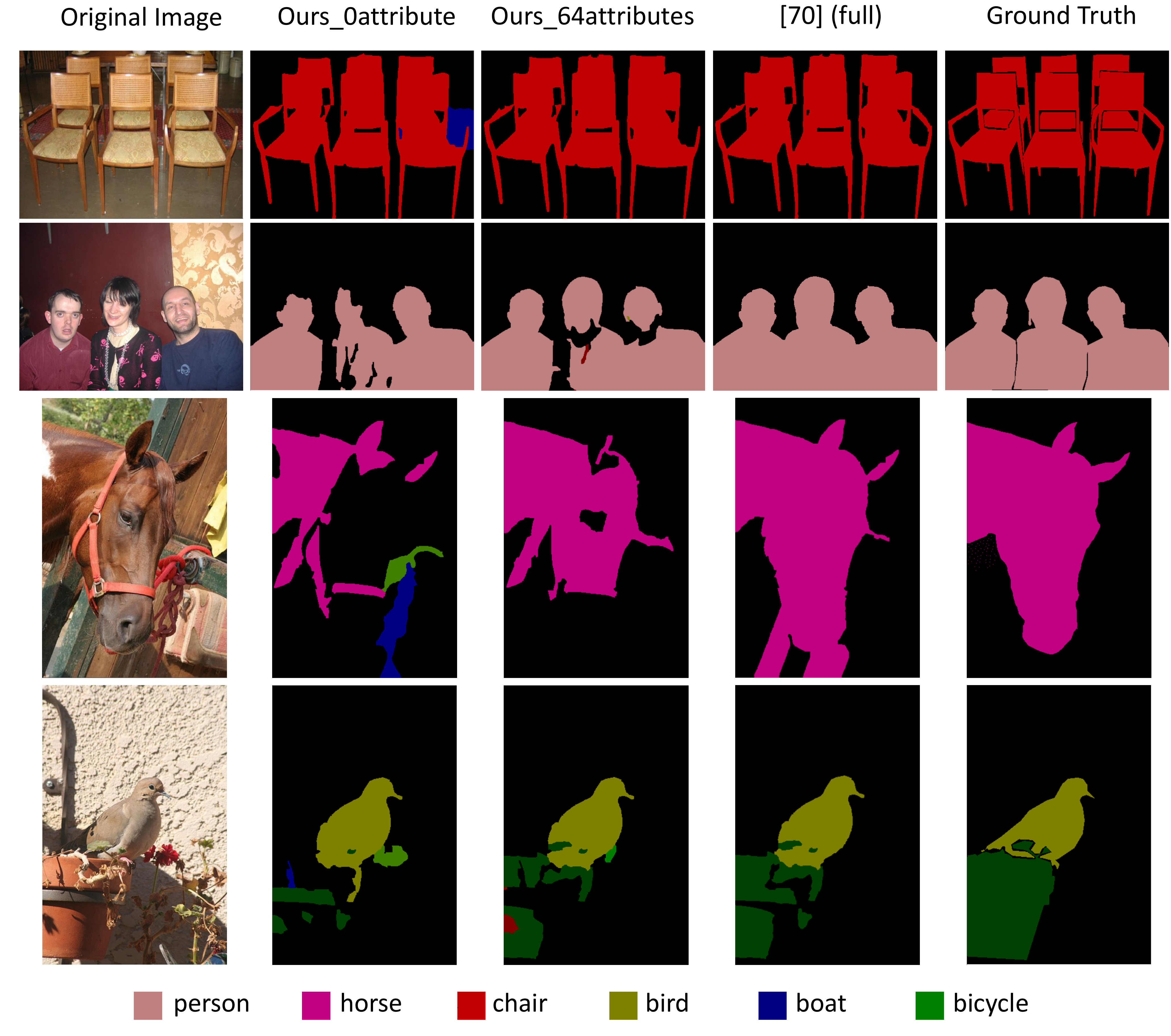}
%\label{1a}
\caption{Qualitative illustration of (attribute-enhanced) semantic segmentation results on aPascal. }
\label{fig:apascal}
\end{figure}

\subsubsection{Results on LabelMe}
\label{sec:labelMe}

\begin{table}[h]
%\scriptsize
%\setlength{\tabcolsep}{0.15em}
%\setlength\extrarowheight{1pt}
\small
\centering
%{\footnotesize
%\begin{tabular}{l || l | l | l ||  l  || l  }
%\scalebox{0.95}{
\begin{tabular}{  M{0.3cm} |  l | c| c  }
\hline
\multicolumn{2}{c |}{Method} & Per-pixel (\%)  & Per-class (\%)   \\
\hline
\parbox{3mm}{\multirow{5}{*}[-4pt]{\rotatebox{90}{Fully}}} & Tighe et al \cite{Tighe_ijcv2013} & 77.0 & 30.1\\
\cline{2-4}
&Tighe et al \cite{Tighe_IJCV2015}  & 78.6 & 39.2\\
\cline{2-4}
&Sigh and Kosecka \cite{singh_cvpr2013}  & 79.2 & 33.8 \\
\cline{2-4}
& Yang et al \cite{Jimei_cvpr2014} & \textbf{79.8} & \textbf{48.7} \\
\cline{2-4}
&Gould et al \cite{Stephen_eccv2014} & 78.4 & 25.7 \\ 
\hline
\hline
\parbox{3mm}{\multirow{7}{*}[-8pt]{\rotatebox{90}{Weakly}}} 
& Vezhnevets et al \cite{vezhnevets_iccv2011} & - & 14 \\
\cline{2-4}
& Vezhnevets et al \cite{vezhnevets_cvpr2012}  & - & 21 \\
\cline{2-4}
& Xu et al \cite{xu_cvpr2014}  & 21.9 & 27.9 \\
\cline{2-4}
& WSDC \cite{Yang_cvpr2013}  & 19.3 & 25.0 \\
\cline{2-4}
& Ours  & \textit{46.2} & \textit{23.8} \\
\cline{2-4}
& Ours\_transductive & \textit{\textbf{52.5}} & \textit{\textbf{31.2}} \\ 
\cline{2-4}
& Ours\_predict  & \textit{48.1} & \textit{26.7} \\
\hline
\end{tabular}%}
%}
\caption{Quantitative comparison of semantic segmentation performance on the LabelMe dataset.}
\label{tab:siftflow}
\end{table}

%\textcolor{blue}{Following the evaluation protocol of fully supervised semantic segmentation approaches \cite{Farabet_tpami2013,Tighe_ijcv2013,Frederick_eccv2014}, we report the overall per-pixel rate which measures the percentage of correctly classified pixels, as well as the average per-class rate which is the percentage of correctly classified pixels for a class. Note that per-pixel rate is dominated by the large classes (e.g. sky, sea, etc.), while per-class rate is dominated by the smaller object classes (e.g. boat, bird).}

Table~\ref{tab:siftflow} compares the performance of our model with a number of state-of-the-art fully supervised \cite{Tighe_ijcv2013,Tighe_IJCV2015,singh_cvpr2013,Jimei_cvpr2014,Stephen_eccv2014} and weakly supervised \cite{vezhnevets_cvpr2012,Yang_cvpr2013,xu_cvpr2014} models\footnote{Very recently the weakly supervised segmentation results have been significantly improved by deep convolutional neural network-based models \cite{Pinheiro_2015_CVPR,Jia_cvpr_2015}. However, these models use much more data to train therefore having an unfair advantage; e.g., the model in \cite{Pinheiro_2015_CVPR} is pre-trained using 1M ImageNet 1K images and fine-tuned  with an additional set of 60K background images.}. Three variants of our models are evaluated: Ours and Ours\_transductive differ in whether the test set images are used for model update (see Sec.~\ref{sec:postproc}), whilst for Ours\_predict, we follow \cite{xu_cvpr2014} and use a pre-trained multi-label image classifier (Linear SVM with ImageNet-trained CNN features as input)  to predict image-level object labels and use those labels for transductive learning. 

The results show that our model outperforms the alternative weakly supervised models \cite{vezhnevets_cvpr2012,Yang_cvpr2013,xu_cvpr2014}  particularly in the per-pixel accuracy which reflects more on the performance on the large classes such as sky and sea. Note that a number of recent weakly supervised learning methods including \cite{Yang_cvpr2013}  are transductive. But our model, even without accessing the whole test set (Ours),  can double the per-pixel accuracy of the alternative models whilst being comparable to them in per-class accuracy. When our model operates in the  transductive mode (Ours\_transductive), the margin over the other  models including \cite{Yang_cvpr2013,xu_cvpr2014} gets even bigger. It is worth mentioning that the result of Xu et al \cite{xu_cvpr2014} is obtained using the predicted image-level labels on the test set. Our result (Ours\_transductive vs. Ours\_predictive) suggests that this additional step is not necessary using our model -- as demonstrated in the image annotation experiments earlier, our model itself can predict image-level labels and does not require assistance from another model.  Table~\ref{tab:siftflow} also shows the performance of a number of state-of-the-art strongly supervised learning models \cite{Tighe_ijcv2013,Tighe_IJCV2015,singh_cvpr2013,Jimei_cvpr2014,Stephen_eccv2014} which require pixel-level annotation of the training images. As can be seen, there is still a fairly big gap between theirs and the best result achieved by our model (Ours\_transductive), although on the per-class metric, it is much closer. %The main reason is that in LabelMe, background stuff such as sky and road are labelled as objects. This limits the scope for our WS-MRF-SIBP to learn latent factors for the background as in the aPascal dataset, thus losing some of the power for disambiguating the weak labels. 
%TMH: This excuse doesn't make sense. We can train from those, we can still use them to disambiguate @ test time.

\begin{figure}[t]
\centering
%\fbox{\includegraphics[height=3.5cm]{ori_A.pdf}}
\includegraphics[width=1.0\linewidth]{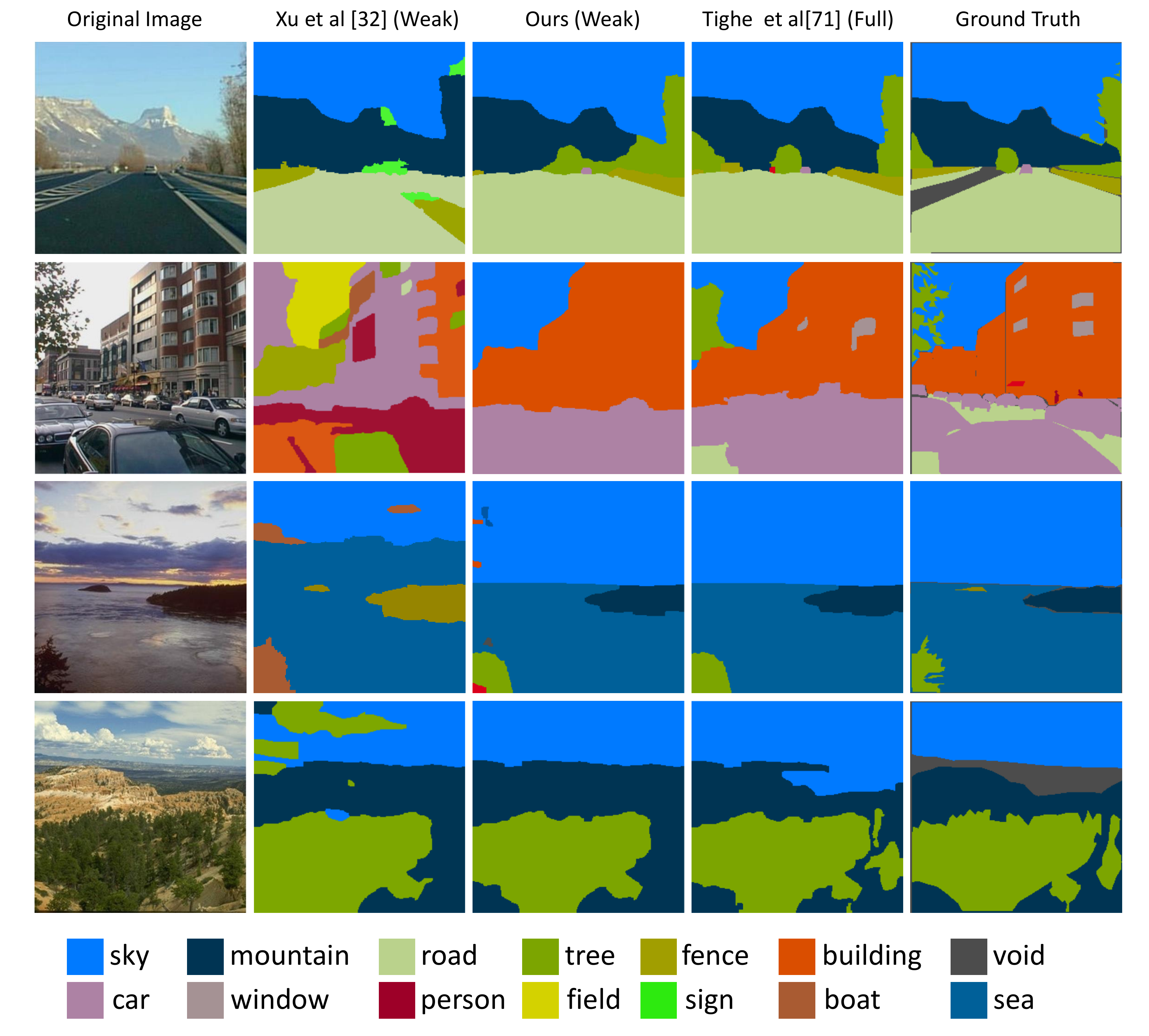}
%\label{1a}
\caption{Qualitative comparison of our semantic segmentation versus alternatives on the LabelMe dataset (best viewed in colour).}
\label{fig:siftflow}
\end{figure}

Fig.~\ref{fig:siftflow} qualitatively compares two weakly supervised methods (\cite{xu_cvpr2014} and Ours) and one fully supervised method (\cite{Tighe_ijcv2013}). We note that the advantage of our model over that in \cite{xu_cvpr2014} is particularly pronounced in the cluttered street scene (second row of Fig.~\ref{fig:siftflow}). For a scene like this, the ability to infer latent factors which correspond to latent attributes for describing object appearance is critical. For example, our model seems to be able to capture the fact that sky can have different types of appearance: clear and blue in the first row, overcast in the second, and cloudy in the bottom. Without accounting for these variations of appearance, the model in \cite{xu_cvpr2014} struggled and  assigned wrong labels to sky in the second and bottom row images.

%\subsection{Further evaluations}
%\label{sec:further eva}

\subsection{Running cost:}

Our unoptimised single-core implementation was run on a PC with an Intel 3.47 GHz CPU and 16GB RAM. The computational time of our approach is comparable to existing methods. Table~\ref{tab:runcost} compares the per iteration time spent on  training. These methods require similar numbers of iterations to converge (around 1500). %The running time is also affected by the number of segments and latent factors $K_{max}$. 
During testing, our method takes about 0.2 seconds per image, comparing to 0.15 using w-LDA. Transductive WSDC does not have a separate train and test stage, and thus there are no directly comparable figures.
% We set $K_{max}=104$ (including 20 objects and 64 attributes) for aPascal dataset.}

\begin{table}[h]
%\footnotesize
%\small
%\begin{tabular*}{5cm}{@{\extracolsep{\fill}}lllr}
\setlength{\tabcolsep}{0.8em}
\centering
{%\footnotesize
%\begin{tabular}{l || l | l | l ||  l  || l  }
%\scalebox{0.95}{
\begin{tabular}{l | l | l | l }

\hline

Method & w-LDA & WSDC \cite{Yang_cvpr2013} & Ours  \\
\hline
Time & 65 & 70 & 80 \\
\hline
\end{tabular}}
%}
\caption{Computation time (seconds per iteration) of different methods on aPascal training set (2113 images).}
\label{tab:runcost}
\end{table}

\section{Conclusion}

We have presented an effective model for weakly-supervised learning of objects, attributes,  and their locations and associations. Learning object-attribute association from weak image-level labels is non-trivial but critical for learning from `natural' data, and scaling to many classes and attributes. We achieve this for the first time through a novel weakly-supervised IBP model that simultaneously disambiguates superpixel annotation correspondence, and learns the appearance of each annotation and superpixel-level annotation correlation. Our results show that on a variety of tasks, our model often performs comparably to strongly supervised alternatives that are significantly more costly to supervise, and is consistently better than weakly supervised alternatives.

The presented model can be improved in a number of ways. First, although the two MRFs integrated in our model capture the spatial label coherence and within-superpixel factorial correlation, \textcolor{black}{other correlations can be considered. One is the image-level correlation capturing object-object co-occurrences in each image, for example, car and road typically co-exist in a street scene. Another is the spatial correlation capturing relative location between objects, for example, sky often being at the top of the road.} Modelling this correlation provides additional constraints to obtain better image and thus superpixel-level labels. However, including this in the current model is non-trivial and a more complex learning algorithm may be needed \cite{velez2009correlatedIBP}. Another prior knowledge the current model does not exploit is the fact that each superpixel should only be explained by a single object label. Although the within-superpixel MRF can implicitly model that, the current model can be be extended to suppress the co-occurrence of multiple object labels explicitly. Finally, the study of automated scene understanding has evolved from single object, multiple object, multiple objects+attributes, towards automated image captioning with full sentences \cite{Vinyals_2015_CVPR,Karpathy_2015_CVPR}, and visual question answering (VQA) \cite{antol2015vqa,hu2016naturalLanguageObjRetrieval}. These  models are discriminative, being hybrids of deep Convolutional Neural Networks (CNNs) and Recurrent Neural Networks (RNNs). Our framework provides a partial answer to captioning -- (it provides a list of objects and their attribute associations) and VQA (it can return the attributes of a queried object, or vice-versa) via an ontology rather than natural language. One future research direction is to integrate to our model with deep learning based language models to tackle the full image captioning/VQA tasks whilst keeping the advantages of the introduced generative non-parametric Bayesian model for weakly supervised learning.

%conclusion can mention handling noise, using other priori, transfer learning etc.
%------------------------------------------------------------------------


%\bibliographystyle{IEEEtran}

%\bibliographystyle{IEEEtranS}
{\small
\bibliographystyle{IEEEtran}
\bibliography{egbib}
}

\vspace{-1cm}
\begin{biography}[{\includegraphics[width=1in,height=1.25in,clip,keepaspectratio]{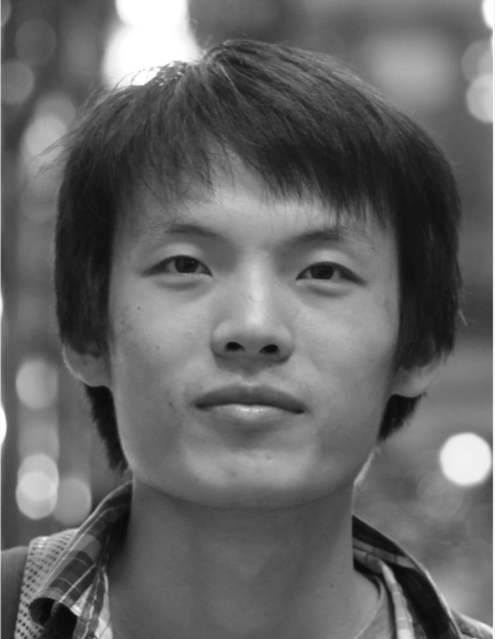}}]{Zhiyuan Shi} received the PhD degree in computer science from Queen Mary University of London in 2016. He is currently postdoctoral associate in computer vision and learning lab at Imperial College London. His research interests include action recognition, deep learning, weakly supervised learning, topic model, object localisation and attribute learning.
\end{biography}
\vspace{-1cm}
\begin{biography}[{\includegraphics[width=1in,height=1.25in,clip,keepaspectratio]{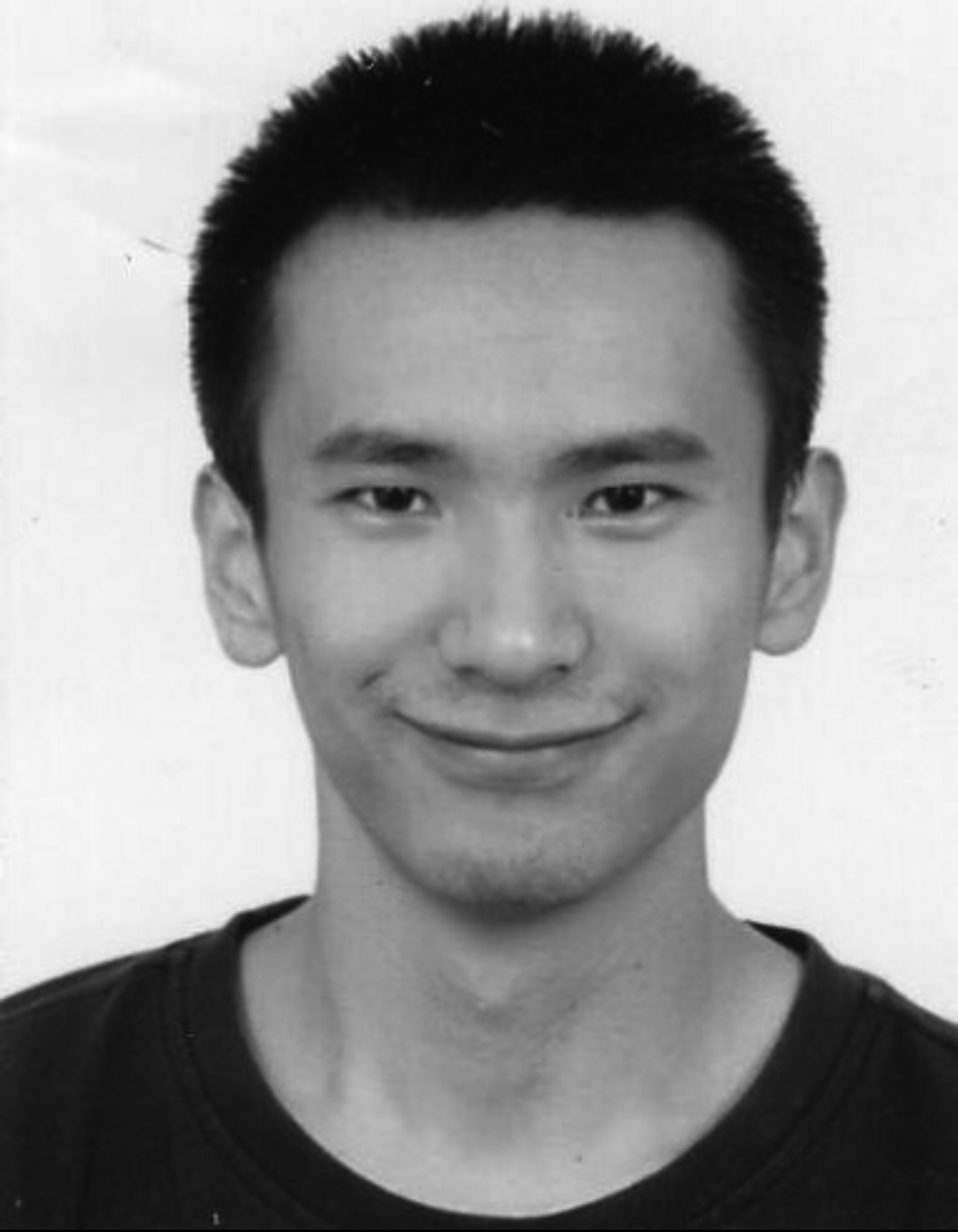}}]{Yongxin Yang} Yongxin Yang is a PhD student at Queen Mary University of London (QMUL). His research interests include machine learning (Transfer Learning, Domain Adaptation, Multi-Task Learning and Deep Learning) and computer vision. 
\end{biography}
\vspace{-1cm}
\begin{biography}[{\includegraphics[width=1in,height=1.25in,clip,keepaspectratio]{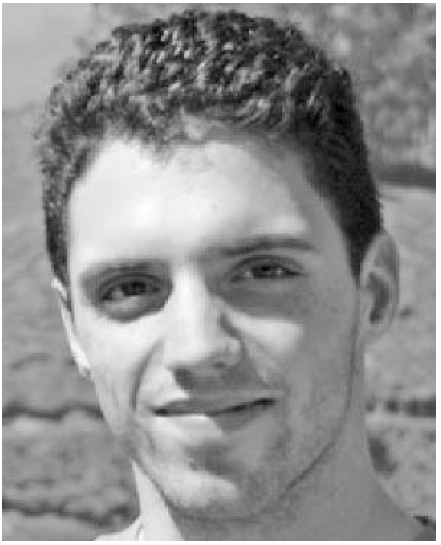}}]{Timothy M. Hospedales} received  the  PhD degree in neuroinformatics from the University of Edinburgh in 2008. He is currently a Reader (associate professor) in Image and Vision Computing at The University of Edinburgh. His research interests include probabilistic modelling and machine learning applied to various problems in computer vision and beyond. He has published more than 60 papers in  major international journals and conferences. He is a member of the IEEE.
\end{biography}
\vspace{-1cm}
\begin{biography}[{\includegraphics[width=1in,height=1.25in,clip,keepaspectratio]{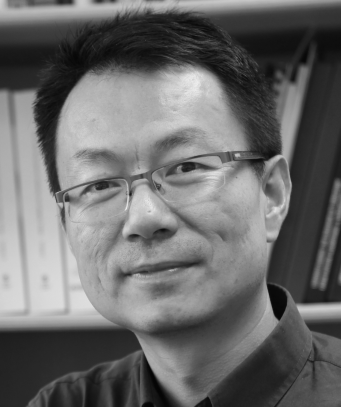}}]{Tao Xiang} received the Ph.D. degree in electrical and computer engineering from the National University of Singapore in 2002. He is currently a reader (associate professor) in the School of Electronic Engineering and Computer Science, Queen Mary University of London. His research interests include computer vision, machine learning, and data
mining. He has published over 120 papers in international journals and conferences.
\end{biography}

\end{document}